\documentclass[sigconf]{acmart}

\renewcommand\footnotetextcopyrightpermission[1]{} 
\AtBeginDocument{%
  \providecommand\BibTeX{{%
    \normalfont B\kern-0.5em{\scshape i\kern-0.25em b}\kern-0.8em\TeX}}}

\usepackage{booktabs,multirow,diagbox,enumitem}
\usepackage{subfig}
\usepackage{cuted}
\usepackage{arydshln}
\usepackage[switch]{lineno}
\usepackage[utf8]{inputenc}
\usepackage{pifont}
\usepackage{graphicx}
\usepackage{caption}
\usepackage{algorithm}
\usepackage{algorithmic}

\newtheorem{definition}{Definition}
\newtheorem{theorem}{Theorem}[section]

\newcommand{\vpara}[1]{\vspace{0.05in}\noindent\textbf{#1 }}
\newcommand{\hide}[1]{} 

\newcommand{\transpose}[1]{\ensuremath{#1^{\scriptscriptstyle T}}}

\begin{document}

\title{Unsupervised Adversarially Robust Representation Learning on Graphs
}

\author{Jiarong Xu}
\affiliation{%
  \institution{Zhejiang University}
  \streetaddress{}
  \postcode{}
}
\email{xujr@zju.edu.cn}

\author{Yang Yang}
\affiliation{%
  \institution{Zhejiang University}
  \streetaddress{}
  \postcode{}
}
\email{yangya@zju.edu.cn}

\author{Junru Chen}
\affiliation{%
  \institution{Zhejiang University}
  \streetaddress{}
  \postcode{}
}
\email{jrchen\_cali@zju.edu.cn}

\author{Chunping Wang}
\affiliation{%
  \institution{FinVolution Group}
  \streetaddress{}
  \postcode{}
}
\email{wangchunping02@xinye.com}

\author{Xin Jiang}
\affiliation{%
  \institution{University of California, Los Angeles}
  \streetaddress{}
  \postcode{}
}
\email{jiangxjames@ucla.edu}

\author{Jiangang Lu}
\affiliation{%
  \institution{Zhejiang University}
  \streetaddress{}
  \postcode{}
}
\email{lujg@zju.edu.cn}

\author{Yizhou Sun}
\affiliation{%
  \institution{University of California, Los Angeles}
  \streetaddress{}
  \postcode{}
}
\email{yzsun@cs.ucla.edu}


\begin{abstract}
Unsupervised/self-supervised pre-training methods for graph representation learning have recently attracted increasing research interests, and they are shown to be able to generalize to various downstream applications. Yet, the adversarial robustness of such pre-trained graph learning models remains largely unexplored. More importantly, most existing defense techniques designed for end-to-end graph representation learning methods require pre-specified label definitions, and thus cannot be directly applied to the pre-training methods. In this paper, we propose an unsupervised defense technique to robustify pre-trained deep graph models, so that the perturbations on the input graph can be successfully identified and blocked before the model is applied to different downstream tasks. Specifically, we introduce a mutual information-based measure, \textit{graph representation vulnerability (GRV)}, to quantify the robustness of graph encoders on the representation space. We then formulate an optimization problem to learn the graph representation by carefully balancing the trade-off between the expressive power and the robustness (\emph{i.e.}, GRV) of the graph encoder. The discrete nature of graph topology and the joint space of graph data make the optimization problem intractable to solve. To handle the above difficulty and to reduce computational expense, we further relax the problem and thus provide an approximate solution. Additionally, we explore a provable connection between the robustness of the unsupervised graph encoder and that of models on downstream tasks. Extensive experiments demonstrate that even without access to labels and tasks, our model is still able to enhance robustness against adversarial attacks on three downstream tasks (node classification, link prediction, and community detection) by an average of +16.5\% compared with existing methods.

\end{abstract}


\maketitle


\section{Introduction} \label{sec:intro}

\begin{figure}[t]  
	\centering  
	\includegraphics[width=0.5\textwidth]{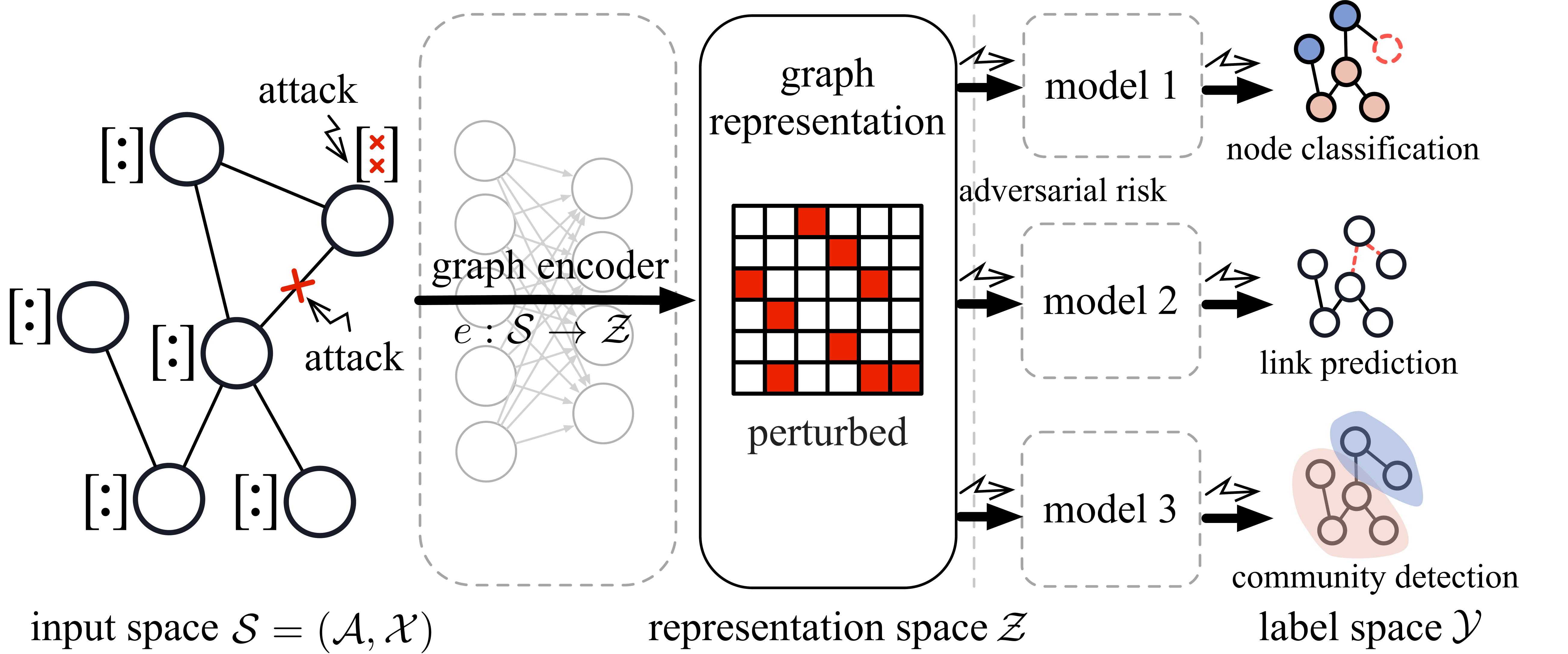}\\
	\caption{Overview of a graph pre-training pipeline under adversarial attacks. 
	If the graph encoder is vulnerable to the attacks, the adversarial risk would propagate to every downstream task via the perturbed graph representation.
	}
	\vspace{-0.08in}
	\label{fig:intro}     
\end{figure}  

Graphs, a common mathematical abstraction for modeling pairwise interactions between objects, are widely applied in numerous domains, including bioinformatics, social networks, chemistry, and finance.
Owing to their prevalence, deep learning on graphs, such as graph neural networks (GNNs)~\cite{kipf2017semi,hamilton2017inductive}, have recently undergone rapid development, making major progress in various analytical tasks, including node classification~\cite{kipf2017semi, hamilton2017inductive}, link prediction~\cite{kipf2016variational}, and graph classification~\cite{xu2018how}. However, most deep learning models on graphs are trained with task-specific labels in an end-to-end manner for a particular task. This motivates some recent efforts to pre-train an expressive graph encoder on unlabeled data and further feed the learned representations to (supervised/unsupervised) off-the-shelf machine learning models for relevant downstream tasks~\cite{hu2019strategies, hu2020gpt, qiu2020gcc}. 
The pre-training models on graphs enable the learned representations to be directly applicable to different applications with a simple and inexpensive machine learning model attached after the encoded representations.

Despite the promising results achieved by deep learning models on graphs, recent studies have shown that these models are vulnerable to adversarial attacks~\cite{dai2018adversarial, zugner2019adversarial, bojchevski2019adversarial}. In other words, even imperceptible perturbations on graph topology and node attributes can significantly affect the learned graph representation, thereby degrading the performance of downstream tasks~\cite{chen2020survey}. This so-called adversarial vulnerability has given rise to tremendous concerns regarding the utilization of deep learning models on graphs, especially in security-critical applications such as drug discovery~\cite{gilmer2017neural} and financial surveillance~\cite{paranjape2017motifs}.
However, the adversarial vulnerability of pre-training models on graphs is far overlooked.
In this work, we show that graph pre-training models also suffer from the adversarial vulnerability problem. 
Actually, owing to the complicated and deep structure, the graph encoder is more vulnerable to adversarial attacks than the simple machine learning models used for downstream tasks in a graph pre-training pipeline~\cite{tanay2016boundary}. 
As Figure~\ref{fig:intro} shows, once the graph encoder is vulnerable to adversarial attacks, the adversarial risk would propagate to every downstream task via the perturbed representations. 

Most efforts targeted on this adversarial vulnerability problem focus on supervised, end-to-end models designed for a particular application scenario~\cite{zugner2019certifiable, bojchevski2019certifiable, wang2019graphdefense, jin2020graph}.
However, the dependency on the supervised information largely limits the scope of their application and usefulness.
For example, these models do not perform well on downstream tasks in which training labels are missing, \emph{e.g.}, community detection in social networks.
In addition, training multiple models for different downstream tasks is both costly and insecure~\cite{feurer5872efficient}.
In contrast, robust unsupervised pre-training models can easily handle the above issues: adversarial attacks are identified and blocked before propagating to downstream tasks.
Moreover, these models are applicable to a more diverse group of applications, including node classification, link prediction, and community detection.
And yet, robust pre-training models under the unsupervised setting remains largely unexplored. 

There are many interesting yet challenging questions in this new field of research. 
Conventionally, the robustness of a model is defined based on the label space~\cite{hao2020adversarial, zugner2019certifiable, bojchevski2019certifiable}, which is not the case in our setting.
Thus the first difficulty we meet is to quantify the robustness of an unsupervised model (without the knowledge of the true or predicted labels).

To overcome the above challenge, in this paper, we first introduce the \emph{graph representation vulnerability} (GRV), an information theoretic-based measure used to quantify the robustness of a graph encoder. We then formulate an optimization problem to study the trade-off between the expressive power of a graph encoder and its robustness to adversarial attacks, measured in GRV.
However, how to efficiently compute or approximate the objective of the optimization problem becomes the next issue.
First, it remains a big problem on how to describe the ability of the attack strategies or the boundary of perturbations, because adversarial attacks on graphs perturb both the discrete graph topology and the continuous node attributes. 
Second, the rigorous definition of the objective 
is intractable.

To handle the above issues, we first quantify the ability of adversarial attacks using Wasserstein distance between probability distributions, and provide a computationally efficient approximation for it. We then adopt a variant of projected gradient descent method to solve the proposed optimization problem efficiently. A sub-optimal solution for the problem gives us a well-qualified, robust graph representation encoder.

Last but not least, we explore several interesting theoretical connections between the proposed measure of robustness (GRV) and the classifier robustness based on the label space.
To show the practical usefulness of the proposed model, we apply the learned representations to three different downstream tasks, namely, node classification, link prediction, and community detection. Experimental results reveal that under adversarial attacks, our model beats the best baseline by an average of +1.8\%, +1.8\%, and +45.8\% on node classification, link prediction, and community detection task, respectively. 

\section{Preliminaries and notations} \label{sec:pre}

In most cases, we use upper-case letters (\emph{e.g.}, $X$ and $Y$) to denote random variables and calligraphic letters (\emph{e.g.}, $\mathcal{X}$ and $\mathcal{Y}$) to denote their support, while the corresponding lower-case letters (\emph{e.g.}, $\boldsymbol{x}$ and $\boldsymbol{y}$) indicate the realizations of these variables. 
We denote the random variables of the probability distributions using  subscripts (\emph{e.g.}, $\mu_{X}$ and $\mu_{Y}$) and the corresponding empirical distributions with hat accents (\emph{e.g.}, $\hat{\mu}_{X}$ and $\hat{\mu}_{Y}$). We use bold upper-case letters to represent matrices (\emph{e.g.}, $\textbf{A}$). When indexing the matrices, $\textbf{A}_{ij}$ denotes the element at the $i$-th row and the $j$-th column, while $\textbf{A}_{i}$ represents the vector at the $i$-th row.
We use $(\mathcal{X}, d)$ to denote the metric space, where $d\, \text{:}\, \mathcal{X} \times \mathcal{X} \rightarrow \mathbb{R}$ is a distance function on $\mathcal{X}$.
We further denote by $\mathcal{M}(\mathcal{X})$ the set of all probability measures on $\mathcal{X}$.

We assume a generic unsupervised graph representation learning setup. In brief, we are provided with an undirected and unweighted graph $\mathbf{G}=(\mathbf{V}, \mathbf{E})$ with the node set $\mathbf{V}= \{v_1, v_2, ..., v_{|\mathbf{V}|}\}$ and edge set $\mathbf{E} \subseteq  \mathbf{V} \times \mathbf{V} = \{e_1, e_2, ..., e_{|\mathbf{E}|}\}$. We are also provided with the adjacency matrix $\mathbf{A} \in \{ 0, 1\}^{|\mathbf{V}| \times |\mathbf{V}|}$ of the graph $\mathbf{G}$, a symmetric matrix with elements $\mathbf{A}_{ij}=1$ if $(v_i, v_j) \in \mathbf{E}$ or $i=j$, and $\mathbf{A}_{ij}=0$ otherwise. We augment $\mathbf{G}$ with the node attribute matrix $\mathbf{X} \in \mathbb{R}^{|\mathbf{V}| \times c}$ if nodes have attributes. Accordingly, we define our input as $\boldsymbol{s} = (\boldsymbol{a}, \boldsymbol{x}) \in \mathcal{S}$; thus, we can conceive of $\boldsymbol{x}$ as the attribute matrix and $\boldsymbol{a}$ as the adjacency matrix of $\mathbf{G}$ under a transductive learning setting, while $\boldsymbol{a}$ and $\boldsymbol{x}$ are the adjacency matrix and attribute matrix respectively of a node's subgraph under an inductive learning setting.
We further define an encoder $e: \mathcal{S} \rightarrow \mathcal{Z}$, which maps an input $\boldsymbol{s} = (\boldsymbol{a}, \boldsymbol{x}) \in \mathcal{S}$ to a representation $e(\boldsymbol{a}, \boldsymbol{x}) \in \mathcal{Z}$, and a simple machine learning model $f: \mathcal{Z} \rightarrow \mathcal{Y} $ that maps a representation $\boldsymbol{z} \in \mathcal{Z}$ to a label $f(\boldsymbol{z}) \in \mathcal{Y}$. 
We go on to define $g = f \circ e$ as their composition, such that $(f \circ e)(\boldsymbol{a}, \boldsymbol{x})=f(e(\boldsymbol{\boldsymbol{a}, \boldsymbol{x}}))$.
A table of main notations is attached in the Appendix.

\vpara{Mutual information.}
Recall that the mutual information between two random variables $X$ and $Y$ is a measure of the mutual dependence between them, and is defined as the Kullback--Leibler (KL) divergence between the joint distribution $p(\boldsymbol x, \boldsymbol y)$ and the product of the marginal distributions $p(\boldsymbol x) p(\boldsymbol y)$:
\begin{align*}
    \mathrm I (X; Y) 
    &= D_\mathrm{KL} \big(p(\boldsymbol x, \boldsymbol y) \| p(\boldsymbol x) p(\boldsymbol y) \big) \\
    &= \int_{\mathcal Y} \int_{\mathcal X} p(\boldsymbol x, \boldsymbol y) \log\left(\frac{p(\boldsymbol x, \boldsymbol y)}{p(\boldsymbol x) p(\boldsymbol y)}\right)  \, dx dy.
\end{align*}
More specifically, it quantifies the ``amount of information'' obtained about one random variable through observing the other random variable.
The successful application of mutual information to various unsupervised graph representation learning tasks has been demonstrated by many authors~\cite{velickovic2018deep, sun2019infograph, tschannen2019mutual}.

\vpara{Admissible perturbations on graphs.}
The Wasserstein distance 
can be conceptualized as an optimal transport problem: we wish to move transport the mass with probability distribution $\mu_S$ into another distribution $\mu_{S^\prime}$ at the minimum cost.
Formally, the $p$-th Wasserstein distance between 
$\mu_S$ and $\mu_{S^\prime}$ 
is defined as
\[
    W_p = (\mu_S, \mu_{S^\prime}) = \left( \inf_{\pi \in \Pi (\mu_S, \mu_{S^\prime})} \int_{\mathcal S^2} d(\boldsymbol s, \boldsymbol s^\prime) \, d\pi(\boldsymbol s, \boldsymbol s^\prime) \right)^{1/p},
\]
where $\Pi(\mu_S, \mu_{S^\prime})$ denotes the collection of all measures on $\mathcal S \times \mathcal S$ with marginal $\mu_S$ and $\mu_{S^\prime}$, respectively.
The choice of $\infty$-Wasserstein distance (\emph{i.e.}, $p=\infty$) is conventional in learning graph representations~\cite{champion2008wasserstein}.

Based on $\infty$-Wasserstein distance, we can quantify the ability of the adversarial attacks.
An attack strategy is viewed as a perturbed probability distribution around that of $S=(A,X)$, and then all possible attack strategies stay in a ball around the genuine distribution~$\mu_S$:
\[
    \mathcal B_\infty(\mu_S, \tau) = \{\mu_{S^\prime} \in \mathcal M(S) \colon W_\infty (\mu_S, \mu_{S^\prime}) \leq \tau\},
\]
where $\tau > 0$ is a pre-defined budget.

\section{Graphs Representations Robust to Adversarial Attacks} \label{sec:model}

In a widely adopted two-phase graph learning pipeline, The first step is to pre-train a graph encoder $e$ (without the knowledge of true labels), which maps the joint input space $\mathcal{S}$ (\emph{i.e.},  the graph topology~$\mathcal A$ and node attributes $\mathcal X$) into some, usually lower-dimensional, representation space $\mathcal{Z}$.
Then the encoded representation is used to solve some target downstream tasks.

In this section, we explain how to obtain a well-qualified graph representation robust to adversarial attacks.
We first propose a measure to quantify the robustness without label information in \S\ref{subsec:rv}.
In~\S\ref{subsec:opt}, we formulate an optimization problem to explore the trade-off between the expressive power and the robustness of the graph encoder.
We then describe every component in the proposed optimization problem, and explain how we obtain a sub-optimal solution efficiently in~\S\ref{subsec:solution}.

\subsection{Quantifying the Robustness of Graph Representations} \label{subsec:rv}

In this section, we propose the \emph{graph representation vulnerability} (GRV) to quantify the robustness of an encoded graph representation.
Intuitively, the learned graph representation is robust if its quality does not deteriorate too much under adversarial attacks.
Now we introduce in detail how to measure the quality of representations using MI, and how to describe the difference of representation quality before and after adversarial attacks.

\vpara{The use of mutual information.} 
A fundamental challenge to achieving a qualified graph representation is the need to find a suitable objective that guides the learning process of the graph encoder. 
In the case of unsupervised graph representation learning, the commonly used objectives are random walk-based~\cite{perozzi2014deepwalk, grover2016node2vec} or reconstruction-based~\cite{kipf2016variational}. 
These objectives impose an inductive bias that neighboring nodes or nodes with similar attributes have similar representations. 
However, the inductive bias is easy to break under adversarial attacks~\cite{jin2020graph, entezari2020all}, leading to the vulnerability of random walk-based and reconstruction-based encoders. 
As an alternative solution, we turn to maximize the MI between the input attributed graph and the representation output by the encoder, \emph{i.e.}, $\mathrm I(S ; e(S))$.
In our case, maximizing the MI $\mathrm I(S; e(S))$ encourages the representations to be maximally informative about the input graph and to avoid the above-mentioned inductive bias.

\vpara{Graph representation vulnerability.}
In addition to the measure of the quality of a graph representation, we also need to describe the robustness of a representation.
Intuitively, an encoder is robust if the MI before and after the attack stay close enough.
Thus, we propose the \emph{graph representation vulnerability (GRV)} to quantify this difference:
\begin{equation} \label{eq:grv}
    \mathrm{GRV}_\tau (e) = \mathrm I (S; e(S)) - \inf_{\mu_{S^\prime} \in \mathcal B(\mu_S, \tau)} \mathrm I(S^\prime; e(S^\prime)),
\end{equation}
where $S=(A,X)$ is the random variable following the benign data distribution, and $S^\prime=(A^\prime, X^\prime)$ follows the adversarial distribution.
The first term $\mathrm I(S;e(S))$ in~\eqref{eq:grv} is the MI between the benign graph data and the encoded representation, while the term $\mathrm I(S^\prime, e(S^\prime))$ uses the graph data after attack.
The attack strategy~$\mu_{S^\star}$ that results in the minimum MI is called the \emph{worst-case} attack, and is defined as
\[
    \mu_{S^\star} = \mathop{\rm argmin}_{\mu_{S^\prime} \in \mathcal B(\mu_S, \tau)} \mathrm I(S^\prime; e(S^\prime)).
\]
Hence by definition, the graph representation vulnerability (GRV) describes the difference of the encoder's behavior using benign data and under the worst-case adversarial attack.
A lower value of $\mathrm{GRV}_\tau(e)$ implies a more robust encoder to adversarial attacks. Formally, an encoder is called $(\tau, \gamma)$-robust if $\mathrm{GRV}_\tau(e) \leq \gamma$.

An analogy to the graph representation vulnerability (GRV) has been studied in the image domain~\cite{Zhu2020LearningAR}. However, the extension of~\cite{Zhu2020LearningAR} to the graph domain requires nontrivial effort. An image is considered to be a single continuous space while a graph is a joint space $\mathcal S=(\mathcal A, \mathcal X)$, consisting of a discrete graph-structure space $\mathcal A$ and a continuous feature space $\mathcal X$. 
Moreover, the perturbations on the joint space $(\mathcal A, \mathcal X)$ is difficult to track because a minor change in the graph topology or node attributes will propagate to other parts of the graph via edges.
This is different in the image domain, where the distributions of all the pixels are assumed to be i.i.d.
Therefore, the discrete nature of graph topology and joint space $(\mathcal A, \mathcal X)$ make the worst-case adversarial attack extremely difficult to estimate. Thus, the optimization method we apply is substantially different from that in~\cite{Zhu2020LearningAR}; see \S\ref{subsec:opt} and \S\ref{subsec:solution}.  Furthermore, more complicated analysis is needed to verify our approach in theory; see \S\ref{sec:theory} for details.

\subsection{Optimization Problem} \label{subsec:opt}

The trade-off between model robustness and the expressive power of encoder has been well-studied by many authors~\cite{tsipras2018robustness, zhang2019theoretically}.
In our case, this trade-off can be readily explored by the following optimization problem
\begin{equation} \label{eq:lb}
    \mbox{maximize} \quad \ell_1(\Theta) = \mathrm I(S;e(S)) - \beta \mathrm{GRV}_\tau (e),
\end{equation}
where the optimization variable is the learnable parameters $\Theta$ of the encoder $e$, and $\beta > 0$ is a pre-defined parameter.

However, in practice, the ``most robust'' encoder is usually not the desired one (as it sacrifices too much in the encoder's expressive power).
An intuitive example for the ``most robust'' encoder is the constant map, which always outputs the same representation whatever the input is.
Hence, a ``robust enough'' encoder would be sufficient, or even better.
To this end, we add a soft-margin $\gamma$ to GRV,
and obtain the following optimization problem
\begin{equation} \label{eq:opt-prob}
    \mathrm{maximize} \quad \ell_2(\Theta) = \mathrm I(S;e(S)) - \beta \max {\{\mathrm{GRV}_\tau(e), \gamma\}}.
\end{equation}
The second term is positive if $\mathrm{GRV}_\tau > \gamma$ and the constant $\gamma$ otherwise.
As a result, when the encoder is sufficiently robust, the second term in~$\ell_2$ vanishes, and thus Problem~\eqref{eq:opt-prob} turns to the standard MI maximization using benign data.
Furthermore, when $\beta=1$, Problem~\eqref{eq:opt-prob} can be divided into two simple sub-problems, depending on the value of $\mathrm{GRV}_\tau(e)$:
\begin{equation} \label{eq:obj}
    \left\{ \begin{aligned}
        &\max_{\Theta} \inf_{\mu_{S^{\prime}} \in \mathcal B_{\infty}(\mu_S, \tau)} \mathrm I (S^\prime; e(S^\prime)),  &&\text{if } \mathrm{GRV}_{\tau} 
        > \gamma \\
        &\max_{\Theta} \quad \mathrm I(S ; e(S)), &&\text{otherwise}.
    \end{aligned} \right.
\end{equation}
In this case ($\beta=1$), when $\mathrm{GRV}_\tau(e) > \gamma$, the problem maximizes the MI under the worst-case adversarial attack.
In other words, the robust encoder tries to maintain the mutual dependence between the graph data and the encoded representation, under all kinds of adversarial attacks. When the encoder is sufficiently robust (\emph{i.e.}, $\mathrm{GRV}_\tau (e) \leq \gamma$), the problem turns to maximize the expressive power of graph encoder.

\vpara{GNN as the parameterized encoder.}
The graph neural network (GNN) has been extensively used as an expressive function for parameterizing the graph encoder~\cite{kipf2016variational, kipf2017semi}.
In this paper, we adopt a one-layer GNN:
$e(\mathbf{A}, \mathbf{X})=\sigma(\hat{\mathbf{D}}^{-1/2} \hat{\mathbf{A}} \hat{\mathbf{D}}^{-1/2} \mathbf{X} \Theta)$,
where $\hat{\mathbf A}$ is the adjacency matrix with self-loops, $\hat{\mathbf D}$ is the corresponding degree matrix, $\sigma$ is the ReLU function, and $\Theta$ is the learnable parameters.

\subsection{Approximate Solution} \label{subsec:solution}

In this section, we discuss in detail how to obtain a sub-optimal solution for Problem~\eqref{eq:obj}.
Overall, the algorithm we use is a variant of the classical gradient-based methods, as presented in Algorithm~\ref{alg:robust}.
In every iteration, we first find out the distribution of the worst-case adversarial attack, and thus we can calculate the value of $\mathrm{GRV}$.
If GRV is larger than $\gamma$, we try to enhance the robustness in this iteration, and apply one gradient descent step for the first sub-problem in Problem~\eqref{eq:obj}.
Otherwise when $\mathrm{GRV}_\tau(e) < \gamma$, the encoder is considered robust enough, and thus we focus on improving the expressive power by one gradient descent move in the second sub-problem in~\eqref{eq:obj}.
As a small clarification, the stopping criterion in Algorithm~\ref{alg:robust} follows the famous early-stopping technique~\cite{prechelt1998early}.

However, there are still many kinds of difficulties in implementing the algorithm.
First of all, the mutual information $\mathrm I(S, e(S))$ is extremely hard to compute, mainly because $S=(A,X)$ is a joint random variable involving a high-dimensional discrete variable $A$.
In addition, the search space of the adversarial attacks, $\mathcal B_\infty (\mu_S, \tau)$, is intractable to quantify:
There is no conventional or well-behaved choice for the distance metric $d$ in such a complicated joint space,
and even when we know the metric, the distance between two random variables is difficult to calculate.
Apart from the above challenges, the classical, well-known projected gradient descent algorithm does not work in the joint space $S=(A,X)$, and thus the worst-case adversarial attack $\mu_{S^\star}$ is no way to find.
Therefore, in this section, we further address the above issues in detail and explain every component in Algorithm~\ref{alg:robust}.

\begin{algorithm}[t]
\caption{Optimization algorithm. 
} \label{alg:robust}
	\begin{flushleft}
		\textbf{Input:} Graph $\mathbf{G}=(\mathbf{A}, \mathbf{X})$, learning rate $\alpha$. \\
		\textbf{Output:} Graph encoder parameters $\Theta$.
	\end{flushleft}
	\begin{algorithmic}[1]
		\STATE Randomly initialize $\Theta$.
		\WHILE {\textit{Stopping condition is not met}}
	        \STATE $\mu_{S^\star} \leftarrow  \mathop{\rm argmin}_{\mu_{S^\prime} \in \mathcal B(\mu_S, \tau)} \mathrm I(S^\prime; e(S^\prime))$.
	    
	        \STATE $\mathrm{GRV}_\tau (e) \leftarrow \mathrm I (S; e(S)) - \mathrm I(S^\star; e(S^\star))$.
		    \IF {$\mathrm{GRV}_\tau(e) > \gamma$}
    		    \STATE $\Theta \leftarrow \Theta - \alpha \nabla_\Theta  \mathrm{I} (S^\star; e(S^\star))$.
		    \ELSE
		        \STATE $\Theta \leftarrow \Theta - \alpha \nabla_\Theta \mathrm I(S; e(S))$.
		    \ENDIF
		\ENDWHILE
	\begin{flushleft}
		\textbf{Return:} $\Theta$.
	\end{flushleft}
	\end{algorithmic}
\end{algorithm}

\vpara{MI estimation.}
Directly computing $\mathrm I(S ; e(S))$ in Problem~\eqref{eq:obj} is intractable, especially for a joint distribution $S=(A,X)$ which includes a high-dimensional discrete random variable $A$.
Some authors propose to maximize the average MI between a high-level ``global'' representation and local regions of the input, and show great improvement in the quality of representations~\cite{hjelm2018learning, velickovic2018deep}.
Inspired by recent work Deep Graph Infomax~\cite{velickovic2018deep}, we use a noise-contrastive type objective as an approximation of the mutual information $\mathrm (I; e(S))$:
\begin{equation} \label{eq:jsd}
\ell_{\text{enc}}(S, e)
= \mathbb{E}_{S}\left[\log \mathcal{D}\left(\boldsymbol{z}, \boldsymbol{z}_{\mathbf{G}}\right)\right]  +  \mathbb{E}_{\tilde{S}}\left[\log \left(1-\mathcal{D}\left(\tilde{\boldsymbol{z}}, \boldsymbol{z}_{\mathbf{G}}\right)\right)\right],
\end{equation}
where $\boldsymbol{z}$ denotes the local representation; $\boldsymbol{z}_{\mathbf{G}} = \text{sigmoid}\left(\mathbb{E}_{S}(\boldsymbol{z})\right)$ represents the global representation; $\tilde{S}$ is the random variable of negative examples, and $\tilde{\boldsymbol{z}}$ represents the realization of $e(\tilde{S})$. 
The critic function $\mathcal{D}(\boldsymbol{z}, \boldsymbol{z}_{\mathbf{G}})$ represents the probability score assigned to a pair of local and global representations obtained from the natural samples (\emph{i.e.}, the original graph), while $\mathcal{D}(\tilde{\boldsymbol{z}}, \boldsymbol{z}_{\mathbf{G}})$ is that obtained from negative samples. 
Common choices for the critic function~$\mathcal{D}$ include bilinear critics, separable critics, concatenated critics, or even inner product critics~\cite{tschannen2019mutual}. 
Here, we select the learnable bilinear critic as our critic function; $\emph{i.e.}$, $\mathcal{D}_{\Phi} = \text{sigmoid}(\transpose{\boldsymbol{z}}\Phi\boldsymbol{z}_{\mathbf{G}})$, where $\Phi$ is a learnable scoring matrix.
Finally, in practice, the expectation over an underlying distribution is typically approximated by the expectation of the empirical distribution over $n$ independent samples $\{(\boldsymbol{a}^i, \boldsymbol{x}^i)\}_{i \in [n]}$.

\vpara{Adversarial distribution estimation.}
Besides the estimation of MI, another challenge involved in solving Problem~\eqref{eq:obj} is how to find the worst-case adversarial distribution $\mu_{S^{\prime}} \in \mathcal{B}_{\infty}(\mu_{S}, \tau)$. 
Here, we divide the difficulties in find $\mu_{S^\star}$ into three categories, and explain in detail how we solve them one by one.

First, it is difficult to choose an appropriate metric $d$ on the joint input space $\mathcal{S}=(\mathcal{A}, \mathcal{X})$ that faithfully measures the distance between each pair of point elements. 
For example, given any pair of points $\boldsymbol s_1 = (\boldsymbol a_1, \boldsymbol x_1)$ and $\boldsymbol s_2 = (\boldsymbol a_2, \boldsymbol x_2)$ in the joint metric space $(\mathcal A, d_\mathcal A)$ and $(\mathcal X, d_\mathcal X)$, an intuitive choice for the distance between $\boldsymbol s_1$ and $\boldsymbol s_2$ would be the $L_p$-norm $\|\big(d_{\mathcal{A}}(\boldsymbol a_1, \boldsymbol a_2), d_{\mathcal X}(\boldsymbol x_1, \boldsymbol x_2) \big)\|_p$.
However, this intuition fails in our case because the changes in graph topology and that in node attributes are not in the same order of magnitude.
Thereby, we have to consider the perturbations in $\mathcal A$ and $\mathcal X$ separately.
With a little abuse of notation, we redefine the perturbation bound as follows:
\begin{align*}
    \mathcal B_{\infty}(\mu_{A}, \mu_{X}, \delta, \epsilon) &= \{(\mu_{A^{\prime}},  \mu_{X^{\prime}}) \in \mathcal{M}(\mathcal{A}) \times \mathcal{M}(\mathcal{X}) \mid \\
    & W_{\infty}(\mu_{A}, \mu_{A^{\prime}}) \le \delta, W_{\infty}(\mu_{X}, \mu_{X^{\prime}}) \le \epsilon \},
\end{align*}
where the small positive numbers $\delta$ and $\epsilon$ play the role of perturbation budget now.
This is indeed a subset of the previous search space $\mathcal B(\mu_S, \tau)$.

Moreover, although the search space has been restricted, the $\infty$-Wasserstein constrained optimization problem remains intractable: We still have no clue about the underlying probability distribution.
Similar to what we did to estimate MI, we turn to replace the real data distribution with an empirical one.
Suppose we have a set of i.i.d. samples $\{(\boldsymbol a^i, \boldsymbol x^i)\}_{i \in [n]}$ (note that $n=1$ under a transductive learning setting, based on which we can compute the empirical distribution $(\hat \mu_A, \hat \mu_X)$).
The empirical search space
is defined as
\begin{align*}
    & \hat{\mathcal B} \big(\{\boldsymbol a^i\}_{i=1}^n, \{\boldsymbol x^i\}_{i=1}^n, \delta, \epsilon\big) \\
    &\qquad = \Big\{(\hat \mu_{A^\prime}, \hat \mu_{X^\prime}) \Bigm\vert \|{\boldsymbol a^i}^\prime - \boldsymbol a^i\|_0 \leq \delta, \|{\boldsymbol x^i}^\prime - \boldsymbol x^i\|_\infty \leq \epsilon, i \in [n] \Big\},
\end{align*}
where $\hat \mu_{A^\prime}$ and $\hat \mu_{X^\prime}$ are the empirical distributions computed from the perturbed samples $\{({{\boldsymbol a}^i}^\prime, {\boldsymbol x^i}^\prime)\}_{i \in [n]}$.
Here we use the cardinality (\emph{i.e.}, $L_0$-norm) to measure the change in graph topology (\emph{i.e.}, $\boldsymbol a$), and the $L_\infty$-norm to measure the change in continuous node attributes (\emph{i.e.}, $\boldsymbol x$).
(When node attributes are discrete, or even binary, we can also use $L_0$-norm for them.)
Finally, we notice that the empirical space $\hat{\mathcal B} \big(\{\boldsymbol a^i\}_{i=1}^n, \{\boldsymbol x^i\}_{i=1}^n, \delta, \epsilon\big)$ is again a subset of $\mathcal B_\infty(\hat \mu_A, \hat \mu_X, \delta, \epsilon)$.

The last, yet the most challenging difficulty is how to efficiently find the worst-case adversarial attack.
We know that he projected gradient descent (PGD) method can be used to find adversarial examples in the image domain~\cite{madry2017towards}.
However, the idea that works well for continuous optimization problems is not directly applicable in our case as the graph topology is a kind of Boolean variables.
Inspired by~\cite{xu2019topology},
as a remedy for the discrete case, we adopt a graph PGD attack for graph topology.
We first find a convex hull of the discrete feasible set, and apply the projected gradient method.
A binary sub-optimal solution $\boldsymbol a^\star$ is then recovered using random sampling.
This variant of PGD helps us identify the worst-case adversarial example efficiently.

\section{Theoretical Connection to Label Space}
\label{sec:theory}

We expect our robust graph encoder is able to block perturbations on graphs and benefits the downstream tasks. To this end, in this section, we establish a provable connection between the robustness of representations (measured by our proposed GRV) and the robustness of the potential model built upon the representations.
Despite the generalization of our framework, we take node classification tasks as an example in this section. 
First, we introduce a conventional robustness, adversarial gap (AG), to measure the robustness of downstream node classifier. Then, we explore some interesting theoretical connections between GRV and AG.

\vpara{Adversarial gap.} 
We here introduce a conventional robustness measure adversarial gap (AG) for node classification, which is built on the label space. 
\begin{definition}
~\label{def:ag}
Suppose we are under inductive learning setting, then $\boldsymbol{a}$ and $\boldsymbol{x}$ are the adjacency matrix and attribute matrix respectively of a node's subgraph.
Let $(\mathcal{S}, d)$ denote the input metric space and $\mathcal{Y} $ be the set of labels. 
For node classification model $g: \mathcal{S} \rightarrow \mathcal{Y} $, we define the \textbf{adversarial risk} of $g$ with the adversarial budget $\tau \geq 0$ as follows:
	\begin{linenomath}
	\begin{equation*}\label{eq:advrisk}
		\begin{array}{ll}
			\text{AdvRisk}_{\tau}(g)
			&= 
			 \mathbb{E}_{{p(\boldsymbol{s}, y)}} 
			[
			\exists\  \boldsymbol{s}^{\prime}=(\boldsymbol{a}^{\prime}, \boldsymbol{x}^{\prime}) \in \mathcal{B}(\boldsymbol{s}, \tau)  \\
			& \text { s.t. } g(\boldsymbol{a}^{\prime}, \boldsymbol{x}^{\prime}) \not= y ]
	    \end{array}
	\end{equation*}
	\end{linenomath}
	Based on $\text{AdvRisk}_{\tau}(g)$, the \textbf{adversarial gap} is defined to measure the relative vulnerability of a given model $g$ w.r.t $\tau$ as follows:
	\begin{linenomath}
	\begin{equation*}
	\text{AG}_{\tau}(g) = 	\text{AdvRisk}_{\tau}(g)  - \text{AdvRisk}_{0}(g).
	\end{equation*}
	\end{linenomath}
\end{definition}
The smaller the value of AdvRisk or AG is, the more robust $g$ is.

\hide{
\begin{definition}
Let $(\mathcal S, d)$ be the input space and $\mathcal Y$ the set of labels.
Suppose $p(s,y^\star)$ is the joint distribution of the input and the ground truth labels.
For any classification model $g \colon \mathcal S \to \mathcal Y$, we define the \textbf{adversarial risk} of $g$ with budget $\tau \geq 0$ as follows
\[
    \mathrm{AdvRisk}_\tau (g) = \mathbb E_{p(s,y^\star)} [s^\prime \mid s^\prime \in \mathcal E],
\]
where the set $\mathcal E$ contains all the adversarial attacks that successfully deceive the classification model $g$, i.e.,
\begin{align*}
    \mathcal E &= \Big\{s^\prime \Bigm\vert d(s^\prime, s) \leq \tau,
    \max_{y \in \mathcal Y \backslash \{y^\star\}} {\big(p(g(s^\prime)=y) - p(g(s^\prime)=y^\star)\big)} > 0 \Big\}.
\end{align*}
\end{definition}
}


\begin{table}[b]
\centering
\caption{Summary of robustness measures. 
\small
Here, the adversarial gap ($\text{AG}$) 
is the robustness measure built on the label space $\mathcal{Y}$, while representation vulnerability ($\text{RV}$) and graph representation vulnerability ($\text{GRV}$) are MI-based measures built on the representation space $\mathcal{Z}$.
The subscript $\epsilon$ denotes the perturbation budget of $\boldsymbol{x}$ (\emph{i.e.}, the image) on the image domain, while the subscript $\tau$ denotes the perturbation budget of $(\boldsymbol{a}, \boldsymbol{x})$ on the graph domain. 
\normalsize
}
\begin{tabular}{p{2cm}<{\centering}p{1cm}<{\centering}p{2cm}<{\centering}p{1cm}<{\centering}}
\toprule
Robustness measure  &Domain & Input space                & Output space               \\\hline
$\text{AG}_{\epsilon}(g)$   &Image & Single $\mathcal{X}$                 & $\mathcal{Y}$ \\
$\text{AG}_{\tau}(g)$      &Graph  & Joint $(\mathcal{A}, \mathcal{X})$   & $\mathcal{Y}$ \\
$\text{RV}_{\epsilon}(e)$  &Image  & Single $\mathcal{X}$                 & $\mathcal{Z}$ \\
$\text{GRV}_{\tau}(e)$      &Graph  & Joint $(\mathcal{A}, \mathcal{X})$   & $\mathcal{Z}$ \\
\bottomrule
\end{tabular}
\label{tab:robust}
\end{table}

Table~\ref{tab:robust} briefly summarize the robustness measures, including AG, RV and GRV.
The traditional model robustness, adversarial gap (\emph{i.e.}, $\text{AG}_{\epsilon}(g)$ and $\text{AG}_{\tau}(g)$), is based on the label space $\mathcal{Y}$, while the MI-based robustness measures (\emph{i.e.}, $\text{RV}_*(e)$ and $\text{GRV}_*(e)$) is built upon the representation space $\mathcal{Z}$.
The prior work~\cite{Zhu2020LearningAR}, which defines $\text{RV}_{\epsilon}(e)$ on a single input space $\mathcal{X}$ in the image domain has shown that $\text{RV}_{\epsilon}(e)$ has a clear connection with classifier robustness.
The graph representation venerability $\text{GRV}_{\tau}(e)$, however, defined as it is on a joint input space $(\mathcal{A}, \mathcal{X})$ in the graph domain, is different from images due to the existence of both discrete and continuous input data structures.
In what follows, we explore some interesting theoretic conclusions that an inherent relationship exists between the graph representation vulnerability  $\text{GRV}_{\tau}(e)$ and the adversarial gap $\text{AG}_{\tau}(g)$; this is based on some certain assumptions that is more aligned with the graph representation learning.

In exploring the GRV's connection to the label space $\mathcal{Y}$, one solution could be to simply assume that one of the input random variables (\emph{i.e.}, $A$ or $X$) and the label random variable $Y$ are independent. 
We first consider the two special cases as follows:
\begin{itemize}[]
\item Topology-aware: given $X \perp Y$, $p(Y|A,X) = p(Y|A)$
\item Attribute-aware: given $A \perp Y$, $p(Y|A,X) = p(Y|X)$
\end{itemize}

Here, we first work on these two special cases under the relevant assumptions. 
We then illustrate a more general case in which $Y$ is dependent on both $A$ and $X$.
Detailed proofs of the following theorems can be found in the Appendix.

\vpara{Special cases.}
We simplify the GNN-based encoder architecture as $\boldsymbol{z} = \transpose{\boldsymbol{a}} \boldsymbol{x}\boldsymbol{\Theta}$ to obtain a tractable surrogate model.
Thus, the representation of each node depends only on its one-hop neighbors, so we can obtain the corresponding column of $\mathbf{A}$ directly to compute the representation for each node. 
Additionally, inspired by~\cite{miyato2016adversarial, AdvT4NE_WWW2019} that have defined the perturbations on the intermediate representations, we opt to define the adversarial distribution w.r.t $\mu_{\transpose{A}X}$ instead of that w.r.t $\mu_{A}$ and $\mu_{X}$ respectively. This assumption is reasonable owing to our focus on the robustness of our model rather than the real attack strategies. Accordingly, we assume that the set of adversarial distributions is $\mathcal{B}_{\infty}(\mu_{\transpose{A}X}, \rho)=\{\mu_{\transpose{{A^{\prime}}}X^{\prime}} \in \mathcal{M}(\mathcal{H}):
W_{\infty}(\mu_{\transpose{A}X}, \mu_{\transpose{{A^{\prime}}}X^{\prime}}) \le \rho \}$  where $\mathcal{H} = \{\transpose{\boldsymbol{a}} \boldsymbol{x}: \forall \boldsymbol{a} \in \mathcal{A}, \boldsymbol{x} \in \mathcal{X} \}$ in the following two theorems.

In Theorems~\ref{thm:att} and~\ref{thm:struct_i}, 
we use $\boldsymbol{a} \in \{0, 1\}^{|\mathbf{V}|}$ to denote a column of $\mathbf{A}$ and $\boldsymbol{x}=\mathbf{X}$. The subscript $\rho$ of GRV, AdvRisk and AG represents that they are defined via $\mathcal{B}_{\infty}(\mu_{\transpose{A}X}, \rho)$, while  $\mathcal{F}=\{f:z \mapsto y \}$ denotes the set of non-trivial downstream classifiers, $f^* = \arg \min_{f \in \mathcal{F}} \text{AdvRisk}_{\rho}(f \circ e)$ is the optimal classifier built upon $e$, and $H_b$ is the binary entropy function. 
Moreover, when indexing $\boldsymbol{a}$ and $\boldsymbol{x}$,  $\boldsymbol{a}_i$ denotes the $i$-th entry of $\boldsymbol{a}$ and $\boldsymbol{x}_i$ denotes the $i$-th row of $\boldsymbol{x}$.

\begin{theorem}[Topology-aware]~\label{thm:att}
Let $(\mathcal{A}, \|\cdot\|_{0})$ and $(\mathcal{X}, \|\cdot\|_{p})$ be the input metric spaces, $\mathcal{Y}=\{-1,+1\}$ be the label space and $\mathcal{Z} = \{-1, +1\}$ be the representation space. The set of encoders with $ \Theta \in \mathbb{R}^{|\mathbf{V}|}$ is as follows:
\begin{align}
\mathcal{E} = \{e: (\boldsymbol{a}, \boldsymbol{x}) \in \mathcal{S} \mapsto \operatorname{sgn}[\transpose{\boldsymbol{a}} \boldsymbol{x}\boldsymbol{\Theta}]|\  \| \boldsymbol{\Theta}\|_2 = 1\}.
~\label{eq:rep1}
\end{align}
Assume that all samples $(\boldsymbol{s}, \boldsymbol{y}) \sim \mu_{SY}$ are generated from $\boldsymbol{y} \stackrel{\text{u.a.r.}}{\sim} U\{-1, +1\}$, $\boldsymbol{a}_i \stackrel{\text{i.i.d.}}{\sim} \text{Bernoulli}(0.5 + \boldsymbol{y}\cdot (p - 0.5))$ and $\boldsymbol{x}_i \stackrel{\text{i.i.d.}}{\sim} \mathcal{N}(\boldsymbol{0}, \sigma^2 \boldsymbol{I}_c)$
where $i=1,2,\dots ,|\boldsymbol{V}|$ and $0 < p < 1$. Then, given $\rho \geq 0$, for any $e \in \mathcal{E}$, we have the following:
\[
    \text{GRV}_{\rho}(e) = 1 - H_b(0.5 + \text{AG}_{\rho}(f^* \circ e)).
\]
Next, consider a simpler case in which $\boldsymbol{y} \stackrel{\text{u.a.r.}}{\sim} U\{-1, +1\}$ and $\boldsymbol{a}_i \stackrel{\text{i.i.d.}}{\sim} \text{Bernoulli}(0.5 + \boldsymbol{y}\cdot (p - 0.5))$ hold, but $\boldsymbol{x}_i = \boldsymbol{1}_c,\ i=1,\dots ,|\boldsymbol{V}|$ and the set of encoders follows such that $\mathcal{E} = \{e : (\boldsymbol{a}, \boldsymbol{x}) \mapsto \operatorname{sgn}[(\transpose{\boldsymbol{a}} \boldsymbol{x} - 0.5|\boldsymbol{V}|\transpose{\boldsymbol{1}}_c) \boldsymbol{\Theta}]\ |\ \| \boldsymbol{\Theta}\|_2 = 1\}$,
which can be regarded as the non-attribute case.
Then, given $\rho \geq 0$, for any $e \in \mathcal{E}$, we have
\begin{linenomath}
\begin{align}
    1 - H_b(0.5 - 0.5 \text{AG}_{\rho}(f^* \circ e)) &\le \text{GRV}_{\rho}(e) \nonumber \\
    \le 1 - H_b(0.5 - &\text{AG}_{\rho}(f^* \circ e)). ~\label{ueq:rv1}
\end{align}
\end{linenomath}
\end{theorem}

Theorem~\ref{thm:att} reveals an explicit connection between  $\text{GRV}_{\rho}(e)$ and $\text{AG}_{\rho}(f^* \circ e)$ achieved by the best classifier in the topology-aware case.
We note that  $H_{b}(\theta)$ is concave on $(0,1)$ and that the maximum of $H_{b}$ is attained uniquely at $\theta=0.5$.
Thus, lower GRV is the sufficient and necessary condition of a smaller AG.

\begin{theorem}[Attribute-aware]~\label{thm:struct_i}
Let $(\mathcal{A}, \|\cdot\|_{0})$ and $(\mathcal{X}, \|\cdot\|_{p})$ be the input metric spaces, $\mathcal{Y}=\{-1,+1\}$ be the label space and $\mathcal{Z} = \{-1, +1\}$ be the representation space. Suppose that the set of encoders is as in ~\eqref{eq:rep1}.
Assume that the samples $(\boldsymbol{s}, \boldsymbol{y}) \sim \mu_{SY}$ are generated from $\boldsymbol{y} \stackrel{\text{u.a.r.}}{\sim} U\{-1, +1\}$, $\boldsymbol{a}_i \stackrel{i.i.d.}{\sim} \text{Bernoulli}(0.5)$ and $\boldsymbol{x}_i \stackrel{i.i.d.}{\sim} \mathcal{N}(\boldsymbol{y}\cdot \boldsymbol{\mu}, \sigma^2 I_c)$
where $i=1,2,\dots ,|\boldsymbol{V}|$. Then, given $\rho \geq 0$, for any $e \in \mathcal{E}$, we have:
\begin{linenomath}
\begin{equation}
    \text{GRV}_{\rho}(e) = 1 - H_b(0.5 - \text{AG}_{\rho}(f^* \circ e)).
\end{equation}
\end{linenomath}
Next, consider a simpler case in which $\boldsymbol{y} \stackrel{\text{u.a.r.}}{\sim} U\{-1, +1\}$, $\boldsymbol{x}_i \stackrel{i.i.d.}{\sim} \mathcal{N}(\boldsymbol{y}\cdot \boldsymbol{\mu}, \sigma^2 I_c)$  but $\boldsymbol{a} \in \{0, 1 \}^{|\boldsymbol{V}|}, \ \sum_{i=1}^{|\boldsymbol{V}|} \boldsymbol{a}_i = n_0 + n_1$, where $n_0 = |\boldsymbol{V}|/4 + \boldsymbol{y}\cdot (p - |\boldsymbol{V}|/4)$, $n_1 = |\boldsymbol{V}|/4 + \boldsymbol{y}\cdot (q - |\boldsymbol{V}|/4)$ and $p+q=|\boldsymbol{V}|/2, \ 0 \le p,q \le |\boldsymbol{V}|/2, \ p,q \in \mathbb{Z}$;
that is, $\transpose{\boldsymbol{a}}\boldsymbol{x}$ will aggregate $n_0$ samples with $\boldsymbol{y}=+1$ and $n_1$ samples with $\boldsymbol{y}=-1$. Further suppose that the set of encoders is as presented in \eqref{eq:rep1}. Then, given $\rho \geq 0$, ~\eqref{ueq:rv1} also holds for any $e \in \mathcal{E}$.
\end{theorem}
Similarly, we have $\text{GRV}_{\rho} \propto \text{AG}_{\rho}$ in  Theorem~\ref{thm:struct_i}.
Note that Theorems~\ref{thm:att} and \ref{thm:struct_i} still hold when $\boldsymbol{a}$ contains self-loops.

\vpara{General case.} 
In the general case, we can extend~\cite[Theorem 3.4]{Zhu2020LearningAR} to the graph domain.
Regardless of the encoder, the theorem below provides a general lower bound of adversarial risk over any downstream classifiers that involves both MI and GRV. We restate the theorem below.
\begin{theorem}\cite{Zhu2020LearningAR}.~\label{thm:gen}
Let $(\mathcal{S}, d)$ be the input metric space, $\mathcal{Z}$ be the representation space and $\mathcal{Y}$ be the label space. Assume that the distribution of labels $\mu_{Y}$ over $\mathcal{Y}$ is uniform and $S$ is the random variable following the joint distribution of inputs $\mu_{AX}$. Further suppose that $\mathcal{F}$ is the set of downstream classifiers. Given $\tau \ge 0$,
\begin{linenomath}
\begin{align*}
    \inf_{f \in \mathcal{F}} \text{AdvRisk}_{\tau}(f \circ e) \ge 1 - \frac{I(S;e(S))-\text{GRV}_{\tau}(e)+\log 2}{\log |\mathcal{Y}|}
\end{align*}
\end{linenomath}
holds for any encoder $e$.
\end{theorem}
Theorem~\ref{thm:gen} suggests that lower adversarial risk over all downstream classifiers cannot be achieved without either lower GRV or higher MI between $S$ and $e(S)$. It turns out that jointly optimizing the objective of maximizing $I(S;e(S))$ and that of minimizing $\text{GRV}_{\tau}(e)$ enables the learning of robust representations. Note that Theorem~\ref{thm:gen} also holds in the graph classification task.

\section{Experiments} \label{sec:exp}
In this section, we demonstrate that our model capable of learning  highly-qualified representations that are robust to adversarial attacks.
In the experiments, we train our model in a fully unsupervised manner, then apply the output representations to three graph learning tasks: namely, node classification, link prediction, and community detection.
We demonstrate that compared with non-robust and other robust graph representation models, the proposed model produces robust representations to defend adversarial attacks (\S~\ref{sec:result-task}).
Furthermore, the superiority of our model still hold under different strengths of attacks (in \S\ref{sec:result-pert}) and under various attach strategies (\S\ref{sec:result-str}).

\begin{table*}[]
	\centering
	\caption{Summary of results for the node classification, link prediction and community detection tasks using polluted data.}
	\vspace{-0.05in}
	\begin{tabular}{l|ccc|ccc|ccc} \toprule
		& \multicolumn{3}{c|}{Node classification (Acc\%)} & \multicolumn{3}{c|}{Link prediction (AUC\%)} & \multicolumn{3}{c}{Community detection (NMI\%)} \\ \midrule
		{\diagbox [width=7em,trim=l] {Model}{Dataset}} &Cora &Citeseer &Polblogs &Cora &Citeseer &Polblogs &Cora &Citeseer &Polblogs \\ \midrule
		$\mathrm{Raw}$ &57.4$\pm$3.0 &49.7$\pm$1.6 &73.9$\pm$0.9 &60.5$\pm$0.1 &50.2$\pm$0.5 &89.0$\pm$0.4 &9.7$\pm$7.5  &1.0$\pm$0.5 &0.2$\pm$0.1\\
		$\mathrm{DeepWalk}$ &56.2$\pm$1.1 &16.5$\pm$0.9 &80.4$\pm$0.5 &55.4$\pm$0.8  &50.3$\pm$0.3  &89.2$\pm$0.7 &34.6$\pm$0.6 &11.1$\pm$1.0 &0.4$\pm$0.5 \\
		$\mathrm{DeepWalk}+\mathbf{X}$ &59.3$\pm$0.4 &26.5$\pm$0.5 &- &55.9$\pm$0.6 &50.9$\pm$0.3 &-  &34.2$\pm$3.7 &11.1$\pm$1.3 &-\\ 
		$\mathrm{GAE}$ &14.0$\pm$1.2 &16.2$\pm$1.1 &49.9$\pm$1.2  &52.4$\pm$1.4  &50.9$\pm$1.8 &50.5$\pm$1.3 &10.9$\pm$2.1  &1.4$\pm$1.7 &9.2$\pm$1.0 \\
		$\mathrm{DGI}$ &69.3$\pm$2.8 &53.2$\pm$2.2 &75.2$\pm$2.4 &68.6$\pm$0.4 &57.6$\pm$2.1 &91.2$\pm$1.1 &30.3$\pm$3.5  &8.5$\pm$3.8 &6.0$\pm$5.6 \\ \midrule
		$\mathrm{Dwns\_AdvT}$ &59.2$\pm$1.2 &25.0$\pm$1.0 &80.7$\pm$0.5 &56.0$\pm$0.7 &50.7$\pm$0.4 &89.5$\pm$0.8 &35.0$\pm$0.7 &11.5$\pm$1.0 &0.9$\pm$0.7\\
		$\mathrm{RSC}$ &46.9$\pm$3.5 &34.0$\pm$2.2 &58.9$\pm$1.7 &52.5$\pm$0.4 &57.2$\pm$0.2 &61.5$\pm$0.4 &4.9$\pm$0.7 &1.8$\pm$0.4 &4.4$\pm$4.3 \\
		$\mathrm{DGI}$-$\mathrm{EdgeDrop}$ &56.0$\pm$4.3 &49.0$\pm$4.5 &79.8$\pm$1.7  &66.2$\pm$0.8 &61.3$\pm$0.9  &89.3$\pm$1.6 &30.1$\pm$6.8  &7.34$\pm$0.8 &9.0$\pm$7.8\\
		$\mathrm{DGI}$-$\mathrm{Jaccard}$ &69.4$\pm$2.8 &57.1$\pm$1.3 &79.3$\pm$0.8 & 63.8$\pm$0.8  &57.6$\pm$1.0 & 84.7$\pm$0.9 &16.4$\pm$1.1  &6.1$\pm$0.6 &12.9$\pm$0.0\\
		$\mathrm{DGI}$-$\mathrm{SVD}$ &68.1$\pm$8.0 &56.1$\pm$16.4 &81.6$\pm$0.7 &60.1$\pm$0.8 &54.7$\pm$1.3 &85.2$\pm$0.7 &16.2$\pm$0.9  &6.5$\pm$0.8 &13.0$\pm$0.0\\ \midrule
		$\mathrm{Ours}$-$\mathrm{soft}$ &69.4$\pm$0.7 &57.5$\pm$2.0 &79.7$\pm$2.1 &68.1$\pm$0.3 &58.2$\pm$1.3 &90.3$\pm$0.5 &39.2$\pm$8.8  &23.5$\pm$1.9 &12.6$\pm$9.6\\
		$\mathrm{Ours}$ &\textbf{70.7$\pm$0.9} &\textbf{58.4}$\pm$1.4 &\textbf{82.7}$\pm$2.2 &\textbf{69.2}$\pm$0.4  &\textbf{59.8}$\pm$1.3 &\textbf{91.8}$\pm$0.4 &\textbf{41.4}$\pm$4.7 &\textbf{23.6}$\pm$2.8 &\textbf{14.8}$\pm$2.7 \\ \bottomrule
	\end{tabular}
	\label{tab:main}
	\vspace{-0.05in}
\end{table*}

\subsection{Experimental Setup}\label{sec:setup}
\vpara{Datasets.}
We conduct experiments on three benchmark datasets: Cora, Citeseer
and  Polblogs.
The first two of these are well-known citation networks  in which nodes are documents and edges are the citation links between two documents. Polblogs is a network of US politics weblogs, where nodes are blogs and the connection between two blogs forms an edge. As Polblogs is a dataset without node attributes, the identity matrix is used to create its node attributes.

\vpara{Baselines.}
The baseline models roughly fall into two categories.
\begin{itemize}[leftmargin=*]
\item \textbf{Non-robust graph representation learning}:
1) Raw: raw features concatenating the graph topology and the node attributes (graph topology (only) for Polblogs);
2) DeepWalk~\cite{perozzi2014deepwalk}: a random walk-based unsupervised representation learning method that only considers the graph topology;
3) DeepWalk+X: concatenating the Deepwalk embedding and the node attributes; 
4) GAE~\cite{kipf2016variational}: variational graph auto-encoder, an unsupervised representation learning method; and
5) DGI~\cite{velickovic2018deep}: another unsupervised representation learning method based on MI.

\item \textbf{Defense models}:
1) Dwns\_AdvT~\cite{10.1145/3308558.3313445}: a defense model designed for Deepwalk; 
2) RSC~\cite{bojchevski2017robust}: a robust unsupervised representation learning method via spectral clustering;
3) DGI-EdgeDrop~\cite{Rong2020DropEdge}: a defense model that works by dropping 10\% of edges during training DGI;
4) DGI-Jaccard~\cite{wu2019adversarial}: DGI applied to a pruned adjacency matrix in which nodes with low Jaccard similarity are forced to be disconnected; and
5) DGI-SVD~\cite{entezari2020all}: DGI applied to a low-rank approximation of the adjacency matrix obtained by truncated SVD.
\end{itemize}
We also include Ours-soft, an variant of our model which removes soft margin on GRV.

\vpara{Implementation details.}
In the training phase, we adopt the graph PGD attack to construct adversarial examples of $\boldsymbol{a}$ while the PGD attack~\cite{madry2017towards} to construct adversarial examples of~$\boldsymbol{x}$.
We set the hyperparameters $\gamma=5\mathrm e\text{-}3$, $\delta=0.4|\mathbf E|$, and $\epsilon=0.1$.
For Polblogs, we do not perform attacks on the constructed node attributes.

In evaluation, we use the same attack strategy as in the training phase. 
Note that DeepWalk and RSC both require the entire graph, 
and thus we have to retrain them using polluted data.
Due to the imperceptible constraint on adversarial attacks, we set $\delta=0.2|\mathbf E|$ during evaluation.
The evaluation is performed on three downstream tasks, and we explain the detailed settings below.
\begin{itemize}[leftmargin=*]
\item Node classification: logistic regression is used for evaluation, and only accuracy score is reported as the test sets are almost balanced. 
For Cora and Citeseer, we use the same dataset splits as in~\cite{kipf2017semi}, but do not utilize the labels in the validation set. For Polblogs, we allocate 10\% of the data for training and 80\% for testing.

\item Link prediction: logistic regression is used to predict whether a link exists or not.
Following conventions,  we generate the positive test set by randomly removing 10\% of existing links and form the negative test set by randomly sampling the same number of nonexistent links. The training set consists of the remaining 90\% of existing links and the same number of additionally sampled nonexistent links. We use the area under the curve (AUC) as the evaluation metric on the link prediction task.

\item Community detection: following the basic schemes for community detection based on graph representation learning, we apply the learned representations to the K-means algorithm. The normalized mutual information (NMI) is used as the evaluation metric here.
\end{itemize}
We run 10 trials for all the experiments and report the average performance and standard deviation.

\subsection{Performance on Downstream Tasks} \label{sec:result-task}

After adversarial attacks on graph topology and node attributes, our model's performance drops by an average of 13.6\%, 1.0\% and 47.3\% on the node classification, link prediction and community detection task, respectively. 
It's worth noting that, in community detection, adversarial attacks can cause dramatic influence on model performance because the community detection task itself is very sensitive to the graph topology.
Table~\ref{tab:main} summarizes the performance of different models in three downstream tasks.
From the table we see that our model beats the best baseline by an average of +1.8\% on the node classification task, +1.8\% on the link prediction task and +45.8\% on the community detection task.
The difference between the performance of our model and that of those non-robust graph learning models indicates the importance of defending adversarial attacks.
Moreover, our model still stands out with huge lead when compared with existing defense models.
Last but not least, the ablation study, \emph{i.e.}, comparing the last two rows in Table~\ref{tab:main}, shows the superiority of the soft margin on GRV.
With this penalty, the model focuses on the representation capability on clean data when the encoder is robust enough (\emph{i.e.}, $\mathrm{RV}_\tau \leq \gamma$), while carefully balances the trade-off between the performance on clean data and the robustness to polluted ones.

\begin{figure}[t]
\centering
	\includegraphics[width=0.5\textwidth]{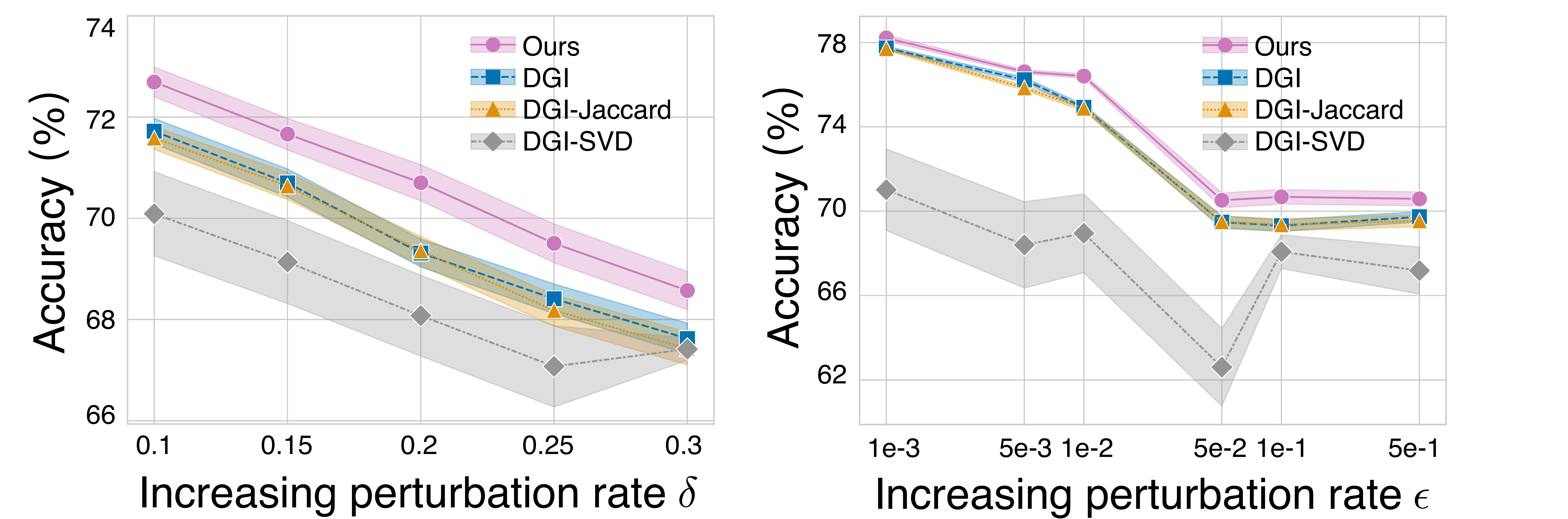}\\
\caption{Accuracy of different models under various perturbation rates $\delta$ and $\epsilon$. The downstream task is node classification and we use the Cora dataset for illustration. The shaded area indicates the standard deviation ($\times 0.1$) over 10 runs.}
\vspace{-0.05in}
\label{fig:robust}
\end{figure} 

\subsection{Performance under Different Rates of Perturbation} \label{sec:result-pert}

We further compare our model with several strong competitors under various perturbation rates.
We use the node classification task and the Cora dataset as an illustrative example.
We vary the strength of adversarial attacks on the graph topology and the node attributes by choosing different perburation rates $\delta$ and $\epsilon$, respectively.
As shown in Figure~\ref{fig:robust}, the performance of our model is consistently superior to other competitors, both on average and in worst-case.
Note that the strong competitor DGI generates negative samples in the training phase,
and this might explain the robustness of the DGI model.
Comparably, the high standard deviation of DGI-SVD might be attributed to the continuous low-rank approximation of the adjacency matrix: the output of truncated SVD is no longer a 0--1 matrix, which violates the discrete nature of graph topology.

\begin{table}[t]	
	\centering
	\caption{Defense against different attackers on Polblogs for the node classification task.}
	\vspace{-0.05in}
	\begin{tabular}{lcccc}
		\toprule
		{\diagbox [width=7em,trim=l] {Model}{Attacker}}  &$\mathrm{Degree}$ &$\mathrm{Betw}$ &$\mathrm{Eigen}$  & $\mathrm{DW}$  \\ \midrule
		$\mathrm{Raw}$ &87.4$\pm$0.3 &84.1$\pm$0.8 &86.4$\pm$0.6 &87.9$\pm$0.4 \\
		$\mathrm{DeepWalk}$ &87.8$\pm$0.9 &83.5$\pm$1.2 &84.3$\pm$1.0 &87.7$\pm$0.9 \\
		$\mathrm{DeepWalk}+\mathbf{X}$ &85.8$\pm$2.7 &82.7$\pm$2.1 &85.0$\pm$1.1 &88.3$\pm$0.9 \\ 
		$\mathrm{GAE}$  &83.7$\pm$0.9 &81.0$\pm$1.6 &81.5$\pm$1.4 &85.4$\pm$1.1 \\
		$\mathrm{DGI}$  &86.6$\pm$1.1 &84.8$\pm$1.2 &84.8$\pm$1.0  &86.4$\pm$1.1    \\
		\hline   
		$\mathrm{Dwns\_AdvT}$ &88.0$\pm$1.0 &84.1$\pm$1.3 &84.6$\pm$1.0 &88.0$\pm$0.8 \\
		$\mathrm{RSC}$  &52.1$\pm$1.3 &51.9$\pm$0.7 &51.4$\pm$0.5 &52.6$\pm$1.1 \\
		$\mathrm{DGI}$-$\mathrm{EdgeDrop}$ &87.1$\pm$0.3 &\textbf{87.0}$\pm$0.6 &80.5$\pm$0.5 &86.3$\pm$0.3 \\
		$\mathrm{DGI}$-$\mathrm{Jaccard}$ &82.1$\pm$0.3 &80.7$\pm$0.4 &80.6$\pm$0.3 &82.2$\pm$0.2 \\
		$\mathrm{DGI}$-$\mathrm{SVD}$ &86.5$\pm$0.2 &85.6$\pm$0.2 &86.1$\pm$0.2 &85.3$\pm$0.3 \\
		\hline
		$\mathrm{Ours}$-$\mathrm{soft}$ &88.5$\pm$0.7 &85.7$\pm$1.5 &86.2$\pm$0.4 &88.7$\pm$0.7 \\
		$\mathrm{Ours}$ &\textbf{89.3}$\pm$0.7 &86.3$\pm$1.2 &\textbf{86.7}$\pm$0.4 &\textbf{89.0}$\pm$0.8  \\
		\bottomrule
	\end{tabular}
	\vspace{-0.08in}
	\label{tab:transfer}
\end{table}

\subsection{Performance under Other Attack Strategies} \label{sec:result-str}

In practice we do not know which kind of attack strategies the malicious users is going to use.
Thus it is interesting and important to know the performance of our model across different types of adversarial attacks.
We adapt some common attack strategies to the unsupervised setting and use them as baselines.
1) Degree: flip edges based on the sum of the degree centrality of two end nodes;
2) Betw: flip edges based on the sum of the betweenness centrality of two end nodes;
3) Eigen: flip edges based on the sum of the eigenvector centrality of two end nodes; and 
4) DW~\cite{bojchevski2019adversarial}: a black-box attack method designed for DeepWalk. We set the size of the sampled candidate set to 20K, as suggested in ~\cite{bojchevski2019adversarial}. 

This time we consider the node classification task on Polblogs dataset for illustration.
This choice is convincing because all the above attack strategies only vary the graph topology, which is the only information we know about Polblogs dataset.
Results in Table~\ref{tab:transfer} shows the outstanding performance of our model as its superiority persists in three attack strategies out of four.
Comparison between Table~\ref{tab:main} and Table~\ref{tab:transfer} shows that the graph PGD attack via MI is the most effective attack strategy used here.
This observation verifies the idea of our model: We learn from the worst adversarial example (\emph{i.e.}, the one deteriorates the performance most).


\section{Related Works}
\label{sec:related}

\vpara{Unsupervised graph representation learning.}
The goal of unsupervised graph representation learning is to learn an encoder that maps the input graph into a low-dimensional representation space.
Currently, the most popular algorithms for unsupervised graph representation learning mainly rely on matrix factorization~\cite{von2007tutorial}, random walks~\cite{perozzi2014deepwalk, grover2016node2vec}, and adjacency matrix reconstruction~\cite{kipf2016variational, duran2017learning}. One alternative approach that has recently been advocated is that of adopting the MI maximization principle~\cite{linsker1988self,sun2019infograph, velickovic2018deep}. This type of methods have achieved massive gain in standard metrics across unsupervised representation learning on graphs, and is even competitive with supervised learning schemes. However, these MI-based graph embeddings usually do not perform well with noisy or adversarial data. This prompts further consideration of MI-based robust representation learning on graphs.

Additionally, there exist some unsupervised graph learning models targeting the adversarial vulnerability problem. One line of them tried to denoise the perturbed input based on certain hypothesis of the crafted attacks~\cite{entezari2020all, wu2019adversarial, dai2018adversarial}, such as the fact that  attackers tend to link two nodes with different features. Another trend focused on the robustness of a particular model, like Deepwalk~\cite{10.1145/3308558.3313445} and spectral clustering~\cite{bojchevski2017robust}.
However, most of them can only operate on certain task, but cannot be applied to various downstream tasks. 

\vpara{Robust models on graphs.}
Owing to the surge of adversarial attacks on graphs, several countermeasures have been proposed. Compared with pre/post-processing approaches, which are supported by empirical observations on specific attacks or models (mostly on GNNs), such as input denoising~\cite{wu2019adversarial, entezari2020all}, adversarial detection~\cite{bojchevski2017robust, xu2018characterizing, ioannidis2019graphsac} and certifiable robustness guarantees on GNNs~\cite{zugner2019certifiable, bojchevski2019certifiable, jin2020certified}, several recent attempts have been made to formulate the graph defense problem as a minimax adversarial game
~\cite{deng2019batch, feng2019graph, xu2019topology, chen2019can, wang2019graphdefense}. 
These approaches often resort to adversarial training for optimization due to its excellent performance; however, they typically require additional label information and are tailored to attacks on GNNs. 
On the other hand, another line of works that utilize a similar adversarial training strategy, do not depend on label information. For instance, one cheap method of this kind involves randomly dropping edges during adversarial training~\cite{dai2018adversarial}.
Moreover, in cases where no label information is exposed,
there are some existing works~\cite{10.1145/3308558.3313445, yu2018learning} that have applied adversarial training to unsupervised representation learning (for example, DeepWalk and autoencoders).
However, the robustness of unsupervised representation learning via MI on graphs remains an inherent blind spot.

The work that most closely resembles ours is that of~\cite{Zhu2020LearningAR}, which develops a notion of representation vulnerability based on the worst-case MI in the image domain. 
However, it cannot address the robustness in the graph domain, mainly caused by 
the joint input space and the discrete nature of graph topology. 
Moreover, the adversarial perturbations of edges or node attributes are easy to propagate to other neighbors via the relational information on a graph, which makes the robustness even harder to enhance.

\vspace{-0.01in}

\section{Conclusion}
In this paper, we study unsupervised adversarially robust representation learning on graphs. 
We propose the graph representation vulnerability (GRV) to quantify the robustness of an unsupervised graph encoder. 
Then we formulate an optimization problem to study the trade-off between the expressive power of the encoder and its robustness to adversarial attacks. 
After that we propose an approximate solution 
which relies on a reduced empirical search space. 
We further build sound theoretical connections between GRV and one example downstream task, node classification.
Extensive experimental results demonstrate the effectiveness of our method on blocking perturbations on input graphs, sregardless of the downstream tasks.


\bibliographystyle{ACM-Reference-Format}

\bibliography{reference}

\appendix

\clearpage

\appendix
\section{Appendix}

\subsection{Notations}
\begin{table}[H]
\centering
\small
	\begin{tabular}{lp{6cm}}
		\toprule 
		\textbf{Notation} & \textbf{Description}   \\
		\midrule
		$\mathbf{G}$, $\mathbf{A}$, $\mathbf{X} $  &  The input graph, the adjacency matrix and the node attribute matrix of $\mathbf{G}$ \\
		$A$, $\boldsymbol{a}$  & The random variable representing structural information and its realization \\
		$X$, $\boldsymbol{x}$  &  The random variable representing attributes and its realization \\
		$S$, $\boldsymbol{s}$ & The random variable $(A, X)$ and its realization $(\boldsymbol{a}, \boldsymbol{x})$ \\
		$\mathcal{A}$, $\mathcal{X}$, $\mathcal{S}$, 
		$\mathcal{Z}$, 
		$\mathcal{Y}$   &  The input space w.r.t graph topology, node attributes, their joint, representations
		and labels. \\
		$\mu_{X}$& The probability distribution of $X$\\
		$\mu_{X^{\prime}}$ & The adversarial probability distribution of $X^{\prime}$ \\
		$\hat{\mu}_{X}^{(n)}$ & The empirical distribution of $X$ \\ $\hat{\mu}_{X^{\prime}}^{(n)}$ & The adversarial empirical distribution of $X^{\prime}$ \\
		$e$, $f$, $g$  &  The encoder function, the classifier function and their composition \\
		\bottomrule 
	\end{tabular}
\normalsize 
\end{table}


\subsection{Implementation Details}
We conduct all experiments on a single machine of Linux system with an Intel Xeon E5 (252GB memory) and a NVIDIA TITAN GPU (12GB memory). 
All models are implemented in PyTorch~\footnote{https://github.com/pytorch/pytorch} version 1.4.0 with CUDA version 10.0 and Python 3.7.

\vpara{Implementations of our model.}
We train our proposed model using the  Adam optimizer  with a learning rate of 1e-3 and adopt early stopping with a patience of 20 epochs. We choose the one-layer GNN as our encoder and set the dimension of its last layer as 512. The weights are initialized via Xavier initialization.

In the training phase, the step size of the graph PGD attack is set to be 20 and the step size of PGD attack is set to be 1e-5. The iteration numbers of both attackers are set to be 10.
In the testing phase, the step size of PGD attack is set to 1e-3. The iteration numbers are set to 50 for both attacks. Others attacker parameters are the same as that in the training phase. 
When evaluating the learned representations via the logistic regression classifier,  we set its learning rate as 1e-2 and train 100 epochs.

\vpara{Implementations of baselines.}
For all the baselines,  we directly adopt their implementations and keep all the hyperparameters as the default values in most cases.
Specifically, for GAE, we adopt their graph variational autoencoder version. Since RSC and $\mathrm{DGI}$-$\mathrm{SVD}$ are models designed for noisy graphs, we grid search their most important hyperparameters
when adopting RSC on benign examples.
For RSC on benign examples, the number of clusters is 128, 200, and 100 on Cora, Citeseer, and Polblogs, respectively.
For $\mathrm{DGI}$-$\mathrm{SVD}$ on benign examples, the rank is set to be 500.

\subsection{Additional Results}

\begin{figure}[t]
\centering
\includegraphics[width=0.18\textwidth]{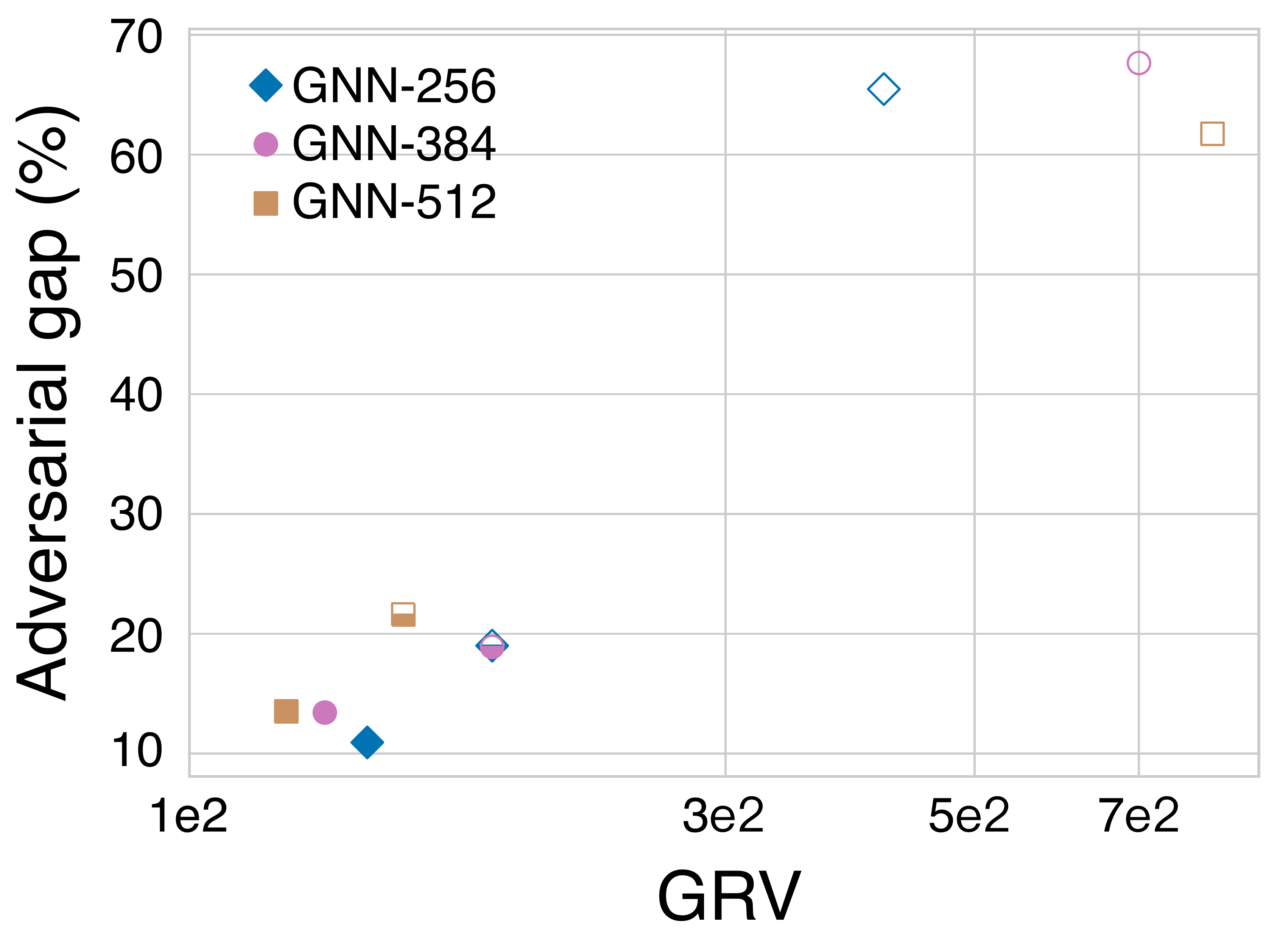}
\hspace{0.01\textwidth}
\includegraphics[width=0.18\textwidth]{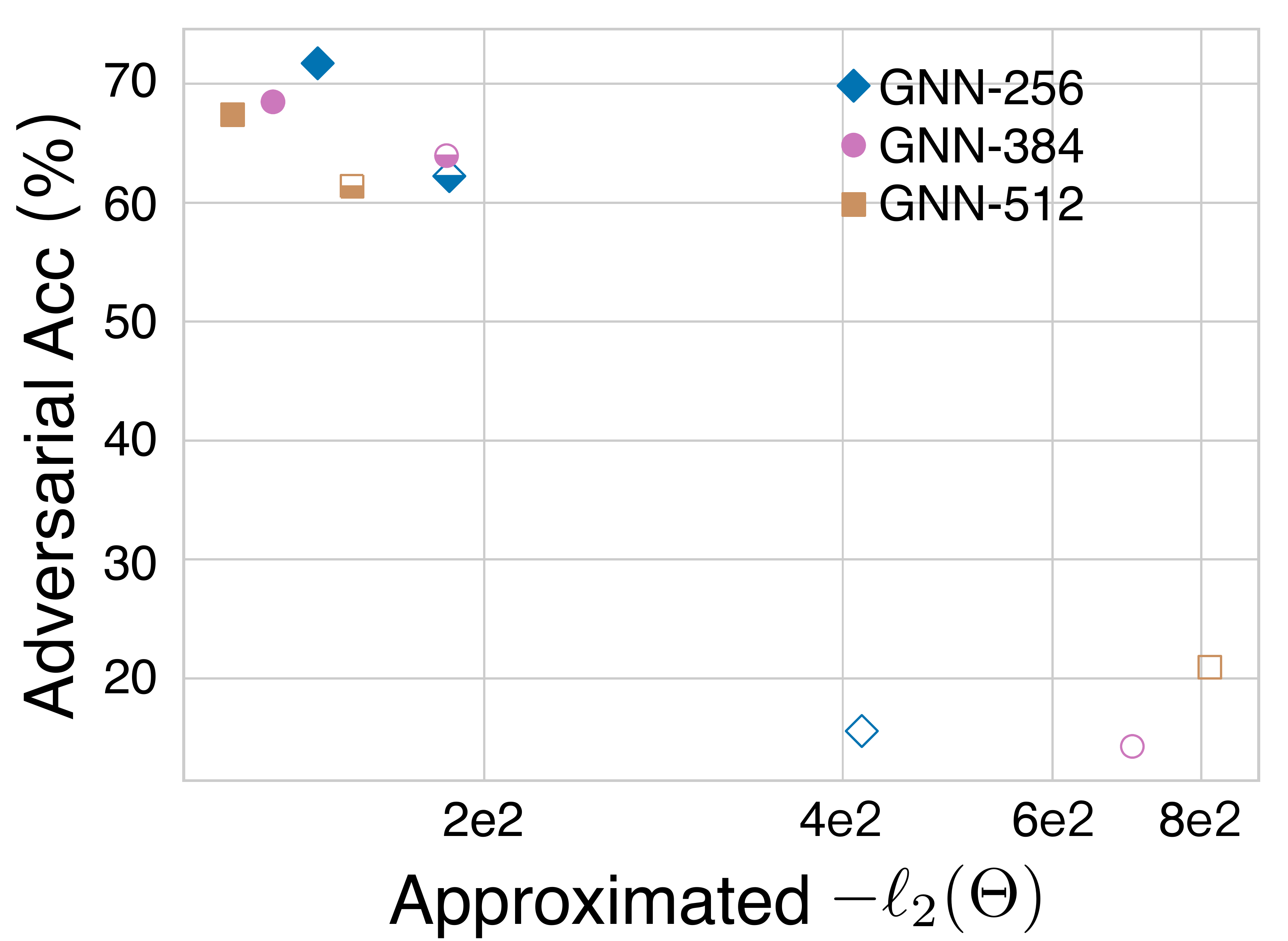}
\caption{\small \emph{Left.} Connection between GRV and AG.
\emph{Right.} Connection between adversarial accuracy and our approximation of $\ell_2(\Theta)$.
Filled points, half-filled points, and unfilled points indicate our models with $\delta=0.4$, $\epsilon=0.1$, our model with $\delta=0.1$, $\epsilon=0.025$, and the DGI model, respectively.
\normalsize 
}
\label{fig:rv}
\end{figure}

\vpara{Empirical Connection between GRV and~AG} \label{sec:result-ag}
In \S\ref{sec:theory}, we established a theoretical connection between the graph representation vulnerability (GRV) and adversarial gap (AG) under different assumptions. Here, we also conduct experiments to corroborate whether a similar connection still holds in more complicated scenarios.
Again, we take the node classification task on the Cora dataset as our illustrative example.
We compare some metrics of three kinds of encoders: GNNs of which the last layers have dimensions 512, 384, and 256, respectively. 
The left of Figure~\ref{fig:rv} presents a positive correlation between the adversarial gap and the value of GRV.
This numerical results shows that GRV is indeed a good indicator for the robustness of graph representations.
Finally, as a supplementary experiment, the right of Figure~\ref{fig:rv} plots the prediction accuracy under polluted data versus our approximation of the objective function $\ell_2(\Theta) = \mathrm I(S; e(S)) - \mathrm{GRV}_\tau (e)$ (with $\beta = 1$).
The figure shows a positive correlation between these two quantities, which verify the use of $\ell_2(\Theta)$ to enhance the adversarial accuracy.

\begin{figure}[]
\begin{minipage}{.49\textwidth}
\centering
\subfloat[Budget of graph topology $\alpha$]{\includegraphics[width=.38\linewidth]{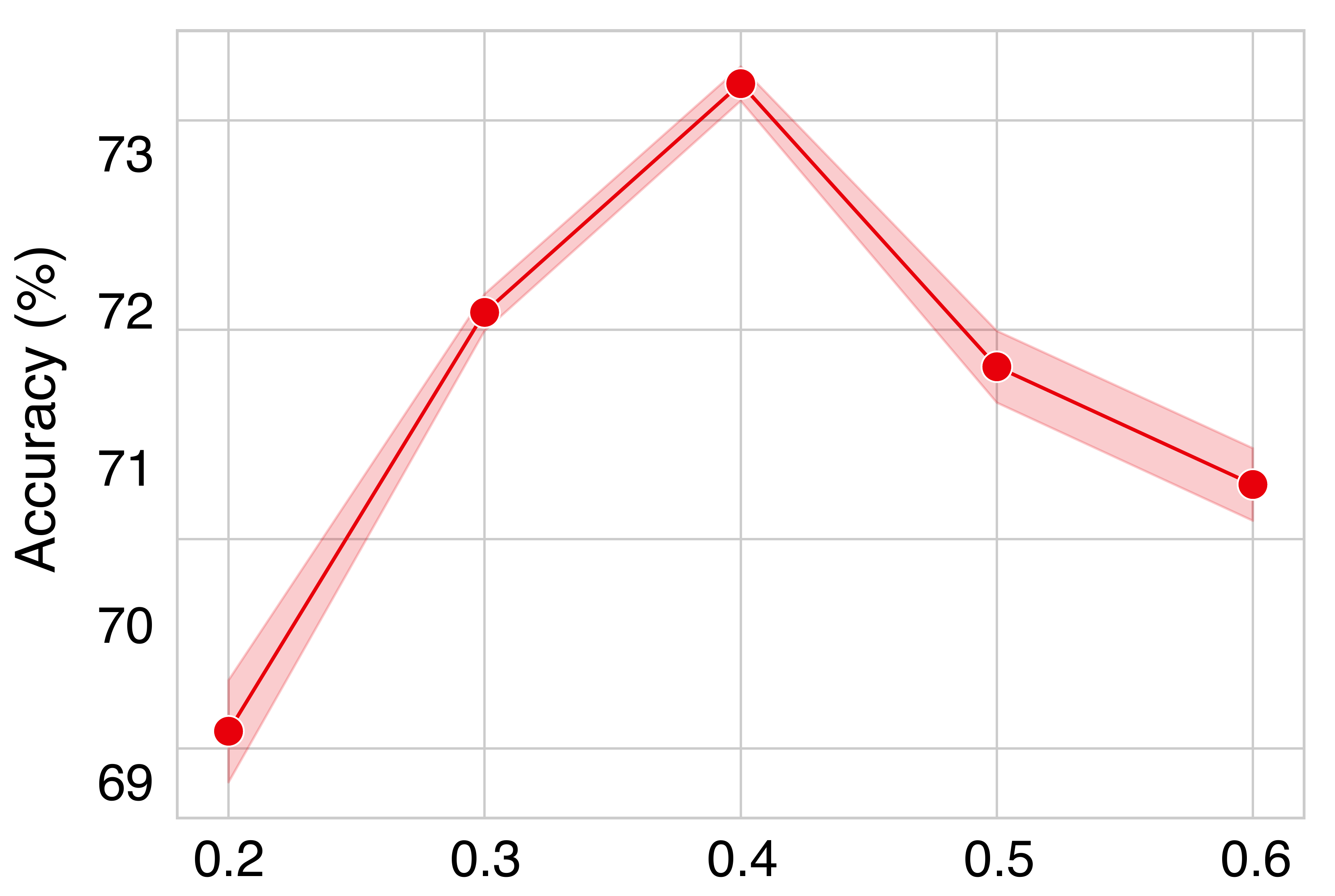}}\quad
\hspace{0.05\textwidth}
\subfloat[Budget of node attributes $\epsilon$]{\includegraphics[width=.38\linewidth]{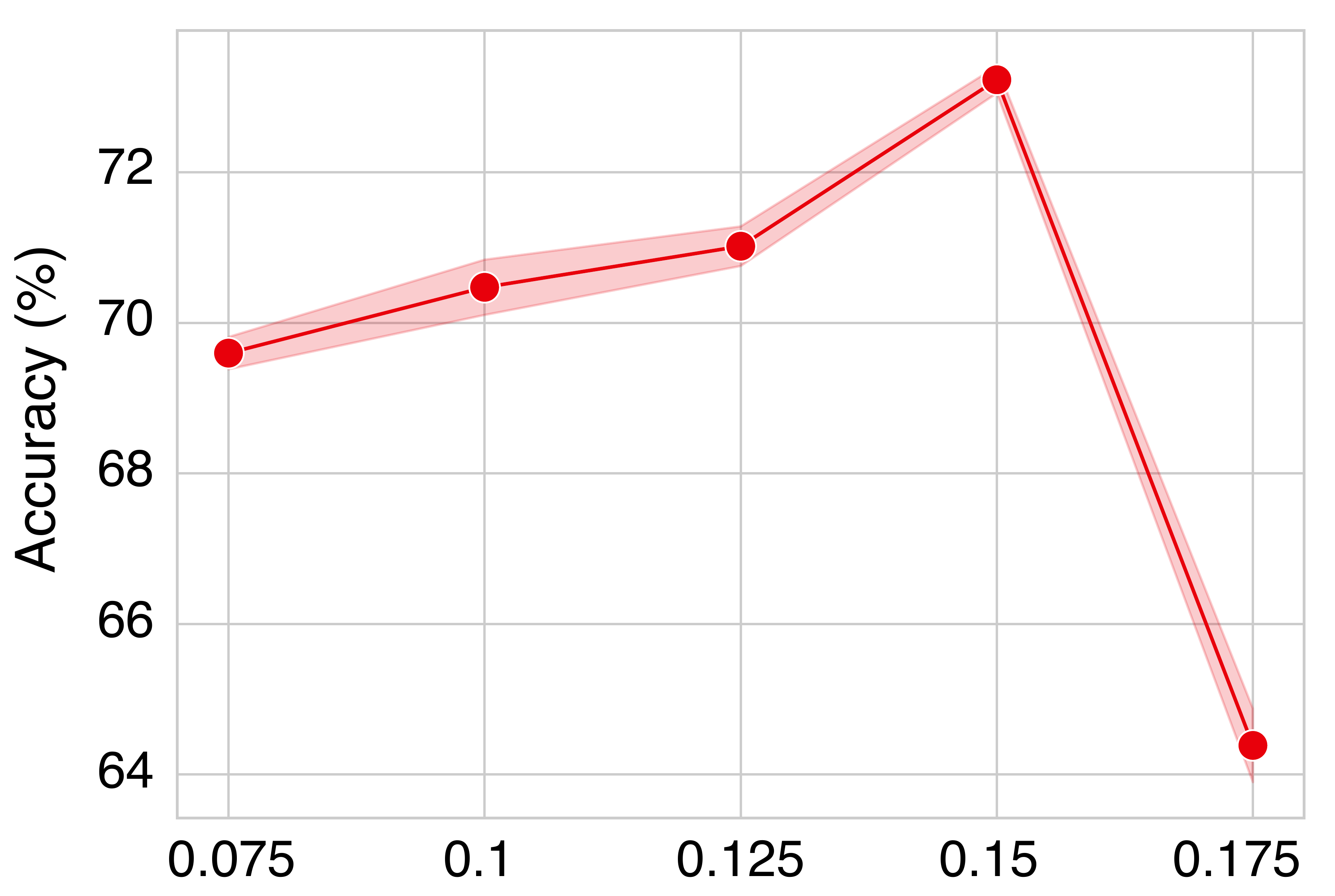}}
\vspace{-0.08in}
\caption{\small Hyperparameter analysis.}
\label{fig:param}
\end{minipage}
\hfill
\begin{minipage}{.49\textwidth}
\centering
\subfloat[Increasing perturbation rate on graph topology]{\includegraphics[width=.38\linewidth]{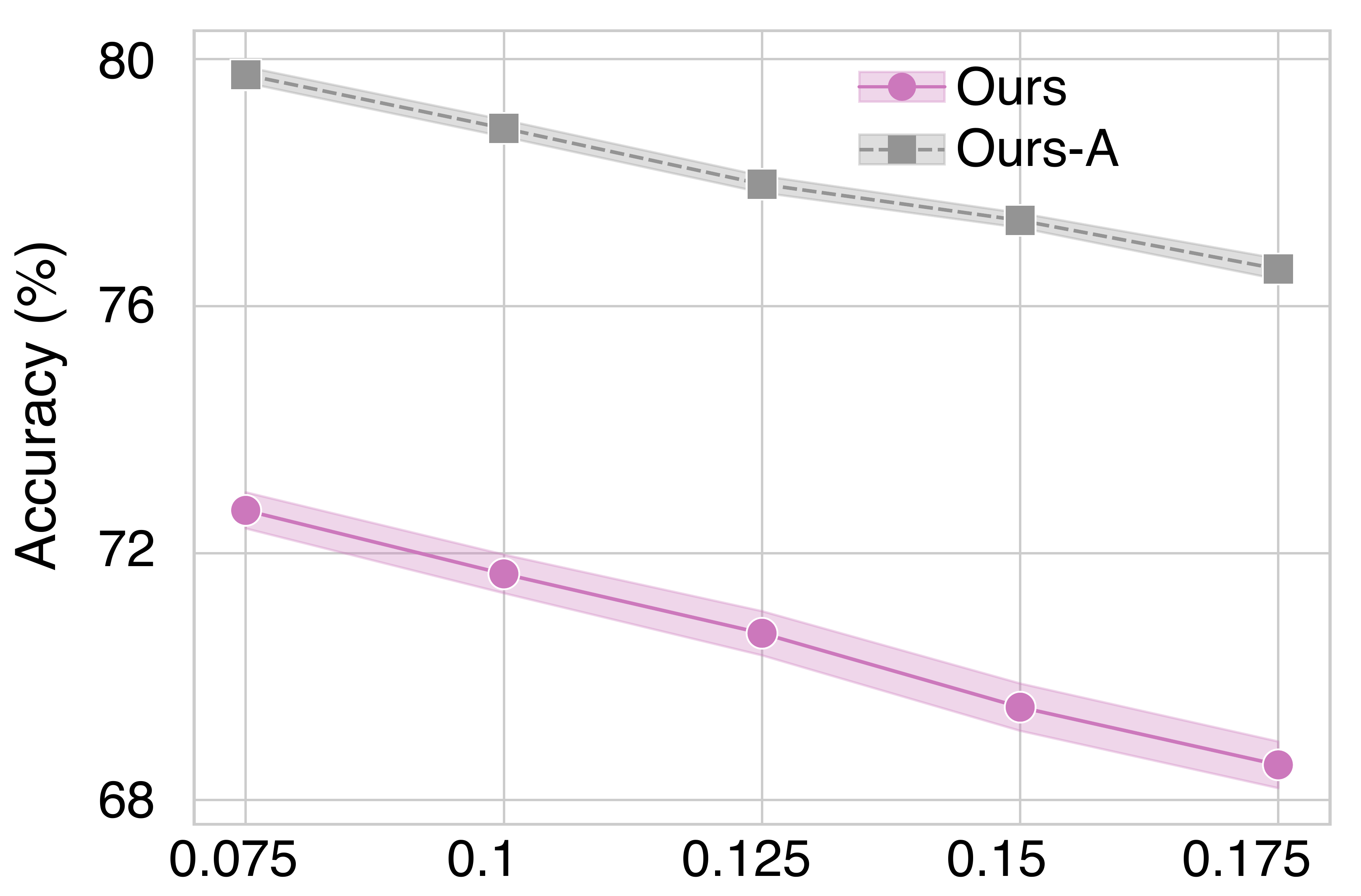}}\quad
\hspace{0.05\textwidth}
\subfloat[Increasing perturbation rate on node attributes]{\includegraphics[width=.38\linewidth]{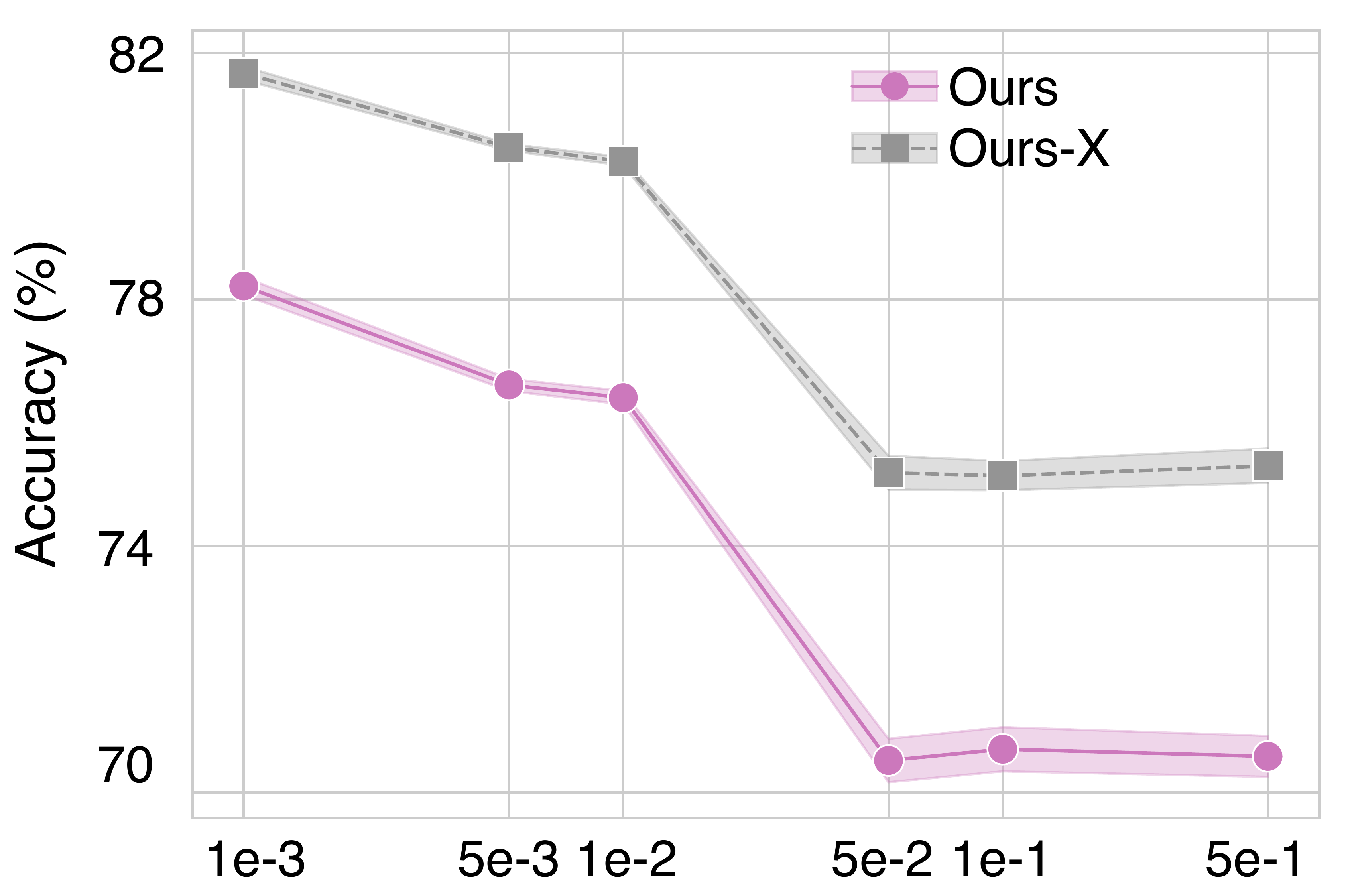}}
\caption{\small The performance of our model against the perturbations performed on a single input space $\mathcal{A}$ or $\mathcal{X}$ compared with that on the joint input space under increasing attack rate.}
\label{fig:AX}
\normalsize 
\end{minipage}
\vspace{-0.08in}
\end{figure}
\vpara{Sensitivity of the budget hyperparameters.}
The budget hyperparameters $\delta$ and $\epsilon$ determine the number of changes made to the original graph topology and node attributes respectively when finding the worst-case adversarial distribution, and are thus important hyperparameters in our proposed model.  We use grid search to find their suitable values in our model
through numerical experiments on Cora in Figure~\ref{fig:param}. Better performance can be obtained when $\alpha=0.3$ and $\epsilon=0.15$. We further observe that when $\alpha$ and $\epsilon$ are  small, the budgets are not sufficient enough to find the worst-case adversarial distribution; while when $\alpha$ and $\epsilon$ are big, introducing too much adversarial attack will also lead to a decrease in the performance to some extent.

\vpara{Defending perturbations on the joint input space is more challenging.}
In all the above-mentioned experiments, we further evaluate our model's robustness against the perturbations performed on the joint space $(\mathcal{A}, \mathcal{X})$. Here, we consider the perturbations performed on a single space $\mathcal{A}$ or $\mathcal{X}$ (\emph{i.e.}, Ours-A/X: our model against perturbations performed on $\mathcal{A}$ or $\mathcal{X}$) under increasing attack rate as the setting of \S~\ref{sec:setup}, and present their results in Figure~\ref{fig:AX}. We can see that when facing with the perturbations on the joint input space, the performance drops 
even more. This indicates that defending perturbations on the joint input space is more challenging, that is why we focus on the model robustness against perturbations on the joint input space.

\subsection{Proofs of Theorems}~\label{subsec:proof}
\setcounter{section}{4}
We only show the proof details of Theorem 4.1. Theorem 4.2 follows in a similar way.

\noindent \textit{\textbf{Proof of Theorem 4.1}.}
For simplicity, we denote $n := |\boldsymbol{V}|$.
Let $R^{\boldsymbol{y}}$ be the random variable following the Binomial distribution according to $A$ (\emph{i.e.}, $R^{\boldsymbol{y}} =\sum_{i=1}^n A_{i} \sim B(n, 0.5+\boldsymbol{y}(p-0.5))$) and  its realization $\boldsymbol{r}^{\boldsymbol{y}}=\sum_{i=1}^n \boldsymbol{a}_{i} $. Let $H$ be the random variable following the Gaussian distribution according to A and X (\emph{i.e.}, $H=\transpose{A}X \sim \mathcal{N}(\boldsymbol{0}, R^{\boldsymbol{y}} \sigma^{2}\boldsymbol{I})$), then its realization $\boldsymbol{h} =\transpose{\boldsymbol{a}} \boldsymbol{x} $  is exactly the aggregation operator of GNNs.
We first compute the explicit formulation of the representation vulnerability $\text{GRV}_{\rho}(e)$. Note that $\mathrm{I}\left(X ; Y\right)=H(X)-H(X|Y)=H(Y)-H(Y|X)$. For any given $e \in \mathcal{E}$, we have
\begin{linenomath}\small
\begin{align*}
\text{GRV}_{\rho}(e) &= \mathrm{I}\left(S ; e\left(S\right)\right) - \inf_{\mu_{\transpose{{A^{\prime}}} X^{\prime}} \sim \mathcal{B}_{\infty}(\mu_{\transpose{A}X}, \rho)}  \mathrm{I}\left(S^{\prime} ; e\left(S^{\prime}\right)\right) \\
&= H_{b}(0.5) - \inf_{\mu_{H^{\prime}} \sim \mathcal{B}_{\infty}(\mu_{H}, \rho)} H_{b}(e(S^{\prime})),
\end{align*}
\normalsize 
\end{linenomath}
because $H(e(S)|S)=0$, and the distribution of $\boldsymbol{h} = \transpose{\boldsymbol{a}} \boldsymbol{x}$,  informally defined as $ \boldsymbol{h} \sim 0.5\mathcal{N}(\boldsymbol{0}, \boldsymbol{r}^{+1}\sigma^{2}\boldsymbol{I}) + 0.5\mathcal{N}(\boldsymbol{0}, \boldsymbol{r}^{-1}\sigma^{2}\boldsymbol{I})$), is symmetric w.r.t. $0$. We thus have, $H_{b}(e(S)) = -P_{\boldsymbol{s} \sim \mu_{S}}(\boldsymbol{h}\Theta \ge 0)\log P_{\boldsymbol{s} \sim \mu_{S}}(\boldsymbol{h}\Theta \ge 0) 
 -P_{\boldsymbol{s} \sim \mu_{S}}(\boldsymbol{h}\Theta < 0)\log P_{\boldsymbol{s} \sim \mu_{S}}(\boldsymbol{h}\Theta < 0) 
= H_{b}(0.5)$.
We note that the binary entropy function $H_{b}(\theta)=-\theta$ $\log(\theta)-(1-\theta)\log(1-\theta)$
is concave on $(0,1)$ and that the maximum of $H_{b}$ is attained uniquely at $\theta=0.5$.
To obtain the infimum of $H_{b}(e(S^{\prime}))$, we should either maximize or minimize $P_{\boldsymbol{s}^{\prime} \sim \mu_{S^{\prime}}}(\boldsymbol{h}^{\prime}\Theta \ge 0)$. 

\vpara{Bound of $P_{\boldsymbol{s}^{\prime} \sim \mu_{S^{\prime}}}(\boldsymbol{h}^{\prime}\Theta \ge 0)$.}
To achieve the bound of $P_{\boldsymbol{s}^{\prime} \sim \mu_{S^{\prime}}}(\boldsymbol{h}^{\prime}\Theta \ge 0)$, we first consider the bound of $|\Delta \boldsymbol{h} \Theta|$ where $\Delta \boldsymbol{h}=\boldsymbol{h}^{\prime}-\boldsymbol{h}=\transpose{(\boldsymbol{a}+\Delta \boldsymbol{a})}(\boldsymbol{x}+\Delta \boldsymbol{x})-\transpose{\boldsymbol{a}}\boldsymbol{x}$.
According to $\mu_{H^{\prime}} \sim \mathcal{B}_{\infty}(\mu_{H}, \rho)$, we can get $\|\Delta H \|_{p} \le \rho$ holds almost surely \emph{w.r.t.} the randomness of $H$ and the transport map defined by $\infty$-Wasserstein distance. 
Then, according to the Hölder’s inequality, we have $|\Delta \boldsymbol{h}\Theta| \le \|\Delta \boldsymbol{h}\|_{p}\|\Theta\|_{q} \le \rho \|\Theta\|_{q}$, which indicates $P_{\boldsymbol{s} \sim \mu_{S}}(|\Delta \boldsymbol{h} \Theta|\le \rho \|\Theta\|_{q})\approx 1$. We have, 
\begin{linenomath}\small
\begin{align*}
    \underbrace{P_{\boldsymbol{s} \sim \mu_{S}}(\boldsymbol{h} \Theta - \rho \|\Theta\|_{q}
    \ge 0)}_{\text{\ding{192}}} &\le P_{\boldsymbol{s}^{\prime} \sim \mu_{S^{\prime}}}(\boldsymbol{h}^{\prime} \Theta \ge 0)  
    \le \underbrace{P_{\boldsymbol{s} \sim \mu_{S}}(\boldsymbol{h} \Theta + \rho \|\Theta\|_{q}\ge 0}_{\text{\ding{193}}} ).
\end{align*}
\normalsize 
\end{linenomath}
\vspace{-0.1in}

\vpara{Compute $\text{GRV}_{\rho}$.}
Next, we will induce the more detailed formulations of the two bounds above. The lower bound is 
\begin{linenomath}\small
\begin{align*}
\text{\ding{192}} 
=& P_{\boldsymbol{h} \sim \mathcal{N}(\boldsymbol{0}, \boldsymbol{r}\sigma^{2}\boldsymbol{I}), \boldsymbol{r} \sim \mu_{R}}(\boldsymbol{h} \Theta - \rho \|\Theta\|_{q} \ge 0) \\
=& P_{Z \sim \mathcal{N}(0, 1)}P_{\boldsymbol{r} \sim \mu_{R}} (Z \ge \rho \|\Theta\|_{q} / \sqrt{\boldsymbol{r}}\sigma\|\Theta\|_{2}).
\end{align*}
\normalsize 
\end{linenomath}
Then according to De Moivre-Laplace Central Limit Theorem, we use Gaussian distribution to approximate Binomial distribution $\boldsymbol{r}^{\boldsymbol{y}}$ ( \emph{e.g.}, $\boldsymbol{r}^{+} \to \mathcal{N}(np,npq)$ where $q=1-p$). We have, 




\begin{linenomath}\small
\begin{align*}
\text{\ding{192}} = &\frac12P_{Z \sim \mathcal{N}(0, 1)}[P_{\boldsymbol{r}^{+} \sim B(n,p)} (Z\sqrt{\boldsymbol{r}^{+}}\sigma\|\Theta\|_{2} \ge \rho \|\Theta\|_{q}) \\
&\qquad\quad\ \ \ \ + P_{\boldsymbol{r}^{-} \sim B(n,1-p)} (Z\sqrt{\boldsymbol{r}^{-}}\sigma\|\Theta\|_{2} \ge \rho \|\Theta\|_{q})] \\
\approx &\frac12 P_{Z \sim \mathcal{N}(0, 1), Z > 0}[P_{Y \sim \mathcal{N}(0,1)} (Y \ge 
M \text{-}  \sqrt{\frac{np}{q}})
+ P_{Y \sim \mathcal{N}(0,1)} (Y \ge
M \text{-}   \sqrt{\frac{nq}{p}})
\\
=:& P_{1}/2 \quad (0 \le P_{1} \le 1).
\end{align*}
\normalsize 
\end{linenomath}
where {\small $M = \frac{\rho^{2}\|\Theta\|_{q}^{2}/Z^{2}\sigma^{2}\|\Theta\|_{2}^{2}}{\sqrt{npq}}$}.
Similarly, we have 
{\small $\text{\ding{193}} \approx  1/2 + P_{2}/2 \quad (0 \le P_{2} \le 1)$}.
Thus, {\small $\text{GRV}_{\rho}(e)=H_{b}(1/2)-H_{b}(\max\{|P_{1}/2-1/2|,|P_{2}/2|\}+1/2)$}.

\vpara{Compute $\text{AG}_{\rho}$.}
Given the formulation of $\text{GRV}_{\rho}$, we further aim to establish its connection to  $\text{AG}_{\rho}$. Here we induce the detailed formulation of $\text{AG}_{\rho}$.
In our case, the only two classifiers to be discussed are $f_{1}(z)=z$ and $f_{2}(z)=-z$. 

For given $e \in \mathcal{E}$, we have $\text{AdvRisk}_{\rho}(f_{1} \circ e)$ as:
\begin{linenomath}\small
\begin{align*}
\text{\ding{194}} :=& \text{AdvRisk}_{\rho}(f_{1} \circ e) =P_{(\boldsymbol{s}, \boldsymbol{y}) \sim \mu_{SY}} 
[\exists\  \boldsymbol{s}^{\prime} \in \mathcal{B}(\boldsymbol{h}, \rho), \text { s.t. } \operatorname{sgn}(\boldsymbol{h}^{\prime}\Theta) \neq \boldsymbol{y}] \\
=& P_{(\boldsymbol{s}, \boldsymbol{y}) \sim \mu_{SY}} 
[\min_{\boldsymbol{s}^{\prime} \in \mathcal{B}(\boldsymbol{h}, \rho)} \boldsymbol{y} \cdot \boldsymbol{h}^{\prime}\Theta \le 0 ] \\
=& P_{(\boldsymbol{s}, \boldsymbol{y}) \sim \mu_{SY}} 
[\boldsymbol{y} \cdot \boldsymbol{h}\Theta \le -\min_{\Delta \boldsymbol{s} \in \mathcal{B}(0, \rho)} \boldsymbol{y} \cdot \Delta\boldsymbol{h}\Theta].
\end{align*}
\normalsize 
\end{linenomath}
Because $|\Delta \boldsymbol{h} \Theta|\le \rho \|\Theta\|_{q}$, $-\min_{\Delta \boldsymbol{s} \in \mathcal{B}(0, \rho)} \boldsymbol{y} \cdot \Delta\boldsymbol{h}\Theta=\rho \|\Theta\|_{q}$ holds for any $\boldsymbol{y}$. We have 
\begin{linenomath}\small
\begin{align*}
\text{\ding{194}} =& \frac12 P_{\boldsymbol{h} \sim \mathcal{N}(\boldsymbol{0}, \boldsymbol{r}^{+} \sigma^{2}\boldsymbol{I}),\boldsymbol{r}^{+} \sim B(n,p)} (\boldsymbol{h} \Theta \le \rho \|\Theta\|_{q}) \\
+& \frac12 P_{\boldsymbol{h} \sim \mathcal{N}(\boldsymbol{0}, \boldsymbol{r}^{-} \sigma^{2}\boldsymbol{I}),\boldsymbol{r}^{+} \sim B(n,q)} (\boldsymbol{h} \Theta \ge -\rho
\approx 1/2 + P_{2}/2.
\end{align*}
\normalsize 
\end{linenomath}
We also have $ \text{AdvRisk}_{\rho=0}(f_{1} \circ e)$ as 
$\text{\ding{195}} := \text{AdvRisk}_{\rho=0}(f_{1} \circ e) = P_{(\boldsymbol{s}, \boldsymbol{y}) \sim \mu_{SY}} 
[ \boldsymbol{y} \cdot \boldsymbol{h}\Theta \le 0 ] = 1/2$.
Thus, $\text{AG}_{\rho}(f_{1} \circ e)=\text{\ding{194}}-\text{\ding{195}}=P_{2}/2$.

Similarly, for given $e \in \mathcal{E}$, we have $\text{AdvRisk}_{\rho}(f_{2} \circ e)$ as $\text{\ding{196}} := \text{AdvRisk}_{\rho}(f_{2} \circ e) \approx 1/2 + P_{2}/2$.
We also have $ \text{AdvRisk}_{\rho=0}(f_{2} \circ e) = 1/2$.
We can get $\text{AG}_{\rho}(f_{2} \circ e)=\text{\ding{196}}-\text{\ding{197}}=P_{2}/2$.

As a result, we have $\text{AG}_{\rho}(f_{1} \circ e)=\text{AG}_{\rho}(f_{2} \circ e)=P_{2}/2$.

\vpara{Connection between GRV and AG.}
Now we aim to find the connection between $\text{AG}_{\rho}$ and $\text{GRV}_{\rho}$. Given their formulations derived above,
It is easy to show that $P_{1}+P_{2}=1$ is equivalent to $1/2-P_{1}/2=P_{2}/2$ and $|P_{1}/2-1/2|=|P_{2}/2|$. Then we have, $\text{GRV}_{\rho}(e) = H_{b}(1/2)-H_{b}(P_{2}/2+1/2) 
        = H_{b}(1/2)-H_{b}(1/2 + \text{AG}_{\rho}(f^{*} \circ e))$,
which completes the proof.

\vpara{Simpler case.}
Consider a simpler case that $\boldsymbol{y} \stackrel{\text{u.a.r.}}{\sim} U\{-1, +1\}$ and $\boldsymbol{a}_i \stackrel{\text{i.i.d.}}{\sim} \text{Bernoulli}(0.5 + \boldsymbol{y} \cdot (p - 0.5))$ hold but $\boldsymbol{x}_i = \boldsymbol{1}_c,\ i=1,2,\dots n,$ and the set of encoders follows that $\mathcal{E} = \{e : (\boldsymbol{a}, \boldsymbol{x}) \mapsto \operatorname{sgn}[(\transpose{\boldsymbol{a}} \boldsymbol{x} - 0.5n\transpose{\boldsymbol{1}}_c) \boldsymbol{\Theta}]\ |\ \| \boldsymbol{\Theta}\|_2 = 1\}$.
This simpler case removes the randomness in $\boldsymbol{x}$. We let 
$H^{\boldsymbol{y}}=\transpose{A}X-0.5n \transpose{\boldsymbol{1}}_c \sim (R^{\boldsymbol{y}}-0.5n)^{c}$, then its realization $\boldsymbol{h} =\transpose{\boldsymbol{a}} \boldsymbol{x} -0.5n \transpose{\boldsymbol{1}}_c $.
We again use the De Moivre-Laplace Central Limit Theorem to approximate $H^{\boldsymbol{y}}$,
\begin{linenomath}\small
\begin{align*}
\boldsymbol{h}^{+} =& \boldsymbol{r}^{+}-0.5n \to \mathcal{N}((np-\frac{n}{2})^{c}, \text{diag}_{c}(npq)), \\
\boldsymbol{h}^{-} =& \boldsymbol{r}^{-}-0.5n \to \mathcal{N}((nq-\frac{n}{2})^{c}, \text{diag}_{c}(npq)).
\end{align*}
\normalsize 
\end{linenomath}
We notice that $np-n/2+nq-n/2=0$ and $H^{\boldsymbol{y}}$ satisfies the conditions of Gaussian Mixture model in \cite{Zhu2020LearningAR}. Thus, we can directly reuse~\cite[Theorem 3.4]{Zhu2020LearningAR} by replacing $\boldsymbol{\theta}^{*}$ with $np-n/2$ and $\boldsymbol{\Sigma}^{*}$ with $\text{diag}_{c}(npq)$, which completes the proof. $\hfill\blacksquare$


\end{document}